%% file: neurips_2026.tex
\documentclass{article}
\usepackage[final,main]{neurips_2026}

\usepackage[utf8]{inputenc}
\usepackage[T1]{fontenc}
\usepackage{hyperref}
\usepackage{url}
\usepackage{xurl}
\usepackage{booktabs}
\usepackage{amsmath}
\usepackage{amsfonts}
\usepackage{nicefrac}
\usepackage{microtype}
\usepackage{xcolor}
\usepackage{graphicx}
\usepackage{multirow}
\usepackage{subcaption}

\newcommand{\rev}[1]{#1}
\newcommand{\fix}[1]{#1}

\title{The Grounding Gap: How LLMs Anchor the Meaning of Abstract Concepts Differently from Humans}

\author{%
  Odysseas S. Chlapanis \\
  Department of Informatics \\
  Athens University of Economics and Business \\
  Archimedes, Athena Research Center \\
  Greece \\
  \And
  Orfeas Menis Mastromichalakis \\
  Instituto de Telecomunica\c{c}\~oes \\
  Portugal \\
  \And
  Christos H. Papadimitriou \\
  Department of Computer Science \\
  Columbia University \\
  United States \\
}

\begin{document}
\maketitle

\begin{abstract}
Abstract concepts --- justice, theory, availability --- have no single perceivable referent; in the human brain, their meaning emerges from a web of experiences, affect, and social context. Do large language models (LLMs) ground abstract concepts in a similar way? We study this by replicating property-generation experiments from cognitive science on 21 frontier and open-weight LLMs. Across models and experiments, we find a consistent pattern: when compared to humans, models rely too heavily on word associations, and underproduce properties tied to emotion and internal states. This yields a large and consistent grounding gap: no model exceeds a Pearson correlation \fix{$r=0.38$} with human responses, compared to a human-to-human ceiling above $r=0.9$. To better interpret this gap, we also replicate a rating experiment on grounding categories and find that here LLMs align more closely with human judgment, and alignment improves as models get larger. We then use sparse autoencoders (SAEs) to inspect whether this information is also reflected in the models' internal features, and we do identify features connected to grounding dimensions such as ``sensorimotor'' and ``social''. These findings suggest that current LLMs can recover grounding dimensions when explicitly queried, but do not recruit them in a human-like way when words are generated freely.\footnote{\fix{Data and code} are publicly available at \url{https://github.com/odychlapanis/grounding-gap/.}}
%

\end{abstract}

\section{Introduction}
\label{sec:intro}
Concrete concepts such as \textit{cat} and \textit{table} are anchored in shared experience  through identifiable referents that can be perceived through the senses \citep{borghi2017challenge}. In contrast, abstract concepts such as \textit{art}, \textit{adventure}, or \textit{justice}, lack such referents.
When humans process abstract words, they reconstruct their meaning from grounded experiences. Consider how a person makes sense of \textit{art}: a colorful \textit{painting}, a feeling of \textit{awe}, a favorite \textit{concert}, as well as more abstract ideas such as \textit{creativity}. Grounded-cognition theories argue that such sensorimotor, emotional, and social anchors are not incidental but constitutive of meaning \citep{lakoff1980metaphors, barsalou1999perceptual, Barsalou2026Grounded}. 
Whether LLMs represent abstract concepts in a similarly grounded way, or rely primarily on patterns of linguistic association \citep{bender-koller-2020-climbing, bender2021dangers}, remains an open question.

To measure how closely LLMs align with humans in the grounding of abstract concepts, we replicate on LLMs two property-generation experiments from cognitive science \citep{harpaintner2018semantic,kelly2024conceptual}. In these experiments, participants --- in these references humans, in this paper models --- are given a concept and asked to list the properties, situations, or associations that come to mind. Although individual responses vary, the coded distributions across human participants are highly consistent: people anchor abstract concepts along grounding dimensions such as sensorimotor experience, emotion, and social context in a similar and consistent way. We then compare the distributions produced by models with those produced by humans. Across 21 frontier and open models, we find a large and consistent grounding gap. In both experiments, \rev{no instruction-tuned model reaches a human--model correlation above $r \approx 0.38$}, while the human--human ceilings exceed $r=0.9$. Models are also much closer to one another than to humans: model--model correlations are systematically higher than model--human correlations. Taken together, these results show that current LLMs exhibit a shared mode of grounding abstract concepts that is systematically different from the human one.

But does the gap arise because models lack an understanding of the categories that structure human grounding, or because this understanding is not recruited in a human-like way when concepts are generated freely? To answer, we turn to a rating experiment from cognitive science, in which words are evaluated directly along grounding-related dimensions \citep{troche2017}. There, recent LLMs align much more closely with human judgments, and this alignment improves with model scale, leaving a gap of only about .1 in Pearson correlation with humans. This contrast suggests that the relevant dimensions are at least partly available to the models when they are explicitly queried, but they are not spontaneously recruited in a human-like way during property generation.

This raises a mechanistic question: Can we find in LLMs internal representations that align with these grounding-related dimensions?
To investigate this, we analyze
models with sparse autoencoders (SAEs), a technique that helps identify interpretable internal features \citep{bricken2023monosemanticity,cunningham2024sparse,templeton2024scaling}. We search for features whose activations track grounding-related dimensions, and \rev{in two Gemma 3 models we find evidence} of such structure: some internal features correlate with human grounding dimensions or sub-dimensions. 
We then validate these features by showing that steering the model along them increases generation of properties from the corresponding target categories. 
Taken together, our findings suggest that current LLMs 
understand the basic grounding categories nearly--but not quite--as well as humans do, and yet they do not recruit them in a human-like way when generating freely.

\paragraph{Contributions.}
\begin{itemize}
    \item We replicate two property-generation experiments from cognitive science on 21 frontier and open LLMs, and show a large and consistent gap between humans and models in the grounding of abstract concepts.

    \item We replicate a rating experiment on grounding-related dimensions and show that LLMs align much more closely -- but not fully -- with human judgments there, indicating that the gap is not simply due to failure to recognize the relevant dimensions.

    \item We use sparse autoencoders to probe \rev{two Gemma 3 models} for internal features related to grounding dimensions, and \rev{find evidence} of such structure.
\end{itemize}

\section{Related work}
\label{sec:related}

\paragraph{Grounded cognition and grounding norms.}
Traditional theories of \emph{grounded cognition} in language posit that abstract categories are not represented as arbitrary symbols but are grounded in sensorimotor and affective experience \citep{barsalou1999perceptual,barsalou2005situating}. Neurocognitive evidence supports that action words engage motor and premotor cortex \citep{hauk2004somatotopic,pulvermuller2005brain}, and emotional and social dimensions of abstract concepts have been linked to interoceptive processing \citep{Critchleyetal2004,mancano2026emotional}. Behavioral norms operationalize these dimensions at scale: datasets in which human participants rate words on Likert scales for dimensions such as sensorimotor strength, arousal, valence, and socialness. \citet{lynott2020lancaster} provide modality-specific strength ratings for 40{,}000 English words, while \citet{scott2019glasgow} and \citet{diveica2023quantifying} cover arousal, valence, and socialness dimensions. These corpora enable grounding research to move beyond binary concrete/abstract distinctions \citep{brysbaert2014concreteness}. The property-listing paradigm \citep{mcrae2005semantic,barsalou2005situating} provides a richer signal: rather than scalar ratings, participants freely generate properties of concepts, yielding a distributional profile over experiential dimensions. \citet{harpaintner2018semantic} applied this paradigm specifically to abstract nouns, showing that they elicit rich sensorimotor, social, and affective features. \citet{kelly2024conceptual} extended the paradigm to test whether emotion concepts form a distinct subcategory, finding that they rely more strongly on interoceptive features than other abstract concepts. These two studies form the empirical basis of our behavioral experiments.

\paragraph{LLM cognitive experiments.}
Early work showed that language models often fail psycholinguistic diagnostics and do not reliably ground semantic properties \citep{ettinger2020bert}. \citet{pezzelle2021word} compared pre-trained transformer representations against human concept norms and found that visual co-training improves alignment for concrete word pairs, but offers little benefit for abstract ones. Recent work extends this question to frontier and open LLMs by prompting them for human-style ratings and comparing them to established norms. \citet{xu2025large} report strong alignment with Glasgow and Lancaster norms on non-sensorimotor dimensions, but weaker alignment on sensory and motor domains. \citet{wang2025cognitive} similarly find persistent gaps for abstract concepts, embodied modalities, and emotion or social dimensions across nine LLMs, with alignment improving with scale. Related studies also show that LLMs can approximate affective and sensorimotor ratings \citep{trott2024can} and recover aspects of perceptual color structure from text alone \citep{abdou2021language}. Closest to our setup, \citet{suresh2023conceptual} use property listing rather than ratings, but focus on concrete concepts and feature-overlap similarity against existing norms. In contrast, we target abstract concepts and classify generated properties into experiential categories.

\paragraph{Mechanistic interpretability of semantic concepts.}
A prominent line of mechanistic interpretability work uses linear probes to localize non-linguistic latent structure in LLM residual streams, recovering concept axes for board game states, spatiotemporal coordinates, and sentiment \citep{li2023emergent,nanda2023emergent,gurnee2024language,tigges2023linear}. Some works extend this paradigm by using these latent vectors to steer models towards the detected concepts \citep{turner2024activation,panickssery2024steering}. Recent studies have identified circuit-level emotion components to control emotional expression \citep{wang2025feel} and recovered emotion directions in models like Claude Sonnet 4.5 \citep{sofroniew2026emotions}. In this work, we analyze features learned by sparse autoencoders (SAEs) to probe whether grounding-related information is reflected in model representations.
SAEs decompose polysemantic LLM representations into monosemantic, interpretable features \citep{bricken2023monosemanticity,cunningham2024sparse,templeton2024scaling}. Recently, SAE features have been applied for analyzing emotion \citep{wu2025ai}. In our work, we leverage the Gemma Scope 2 SAE models \citep{gemmascope2_2025}, the successor to Gemma Scope \citep{lieberum-etal-2024-gemma}, and we probe sensorimotor, internal-state, and social concepts.

\section{The grounding gap}

\label{sec:experiments}

We study LLM grounding of abstract concepts through \emph{property generation}, a behavioral paradigm from cognitive science in which participants are given a concept and asked to list the features, situations, or associations that come to mind for it. Unlike scalar rating tasks \citep{troche2017}, which require participants to map conceptual content onto a fixed set of predefined dimensions, property generation captures the content that people spontaneously recruit when representing a concept. For this reason, it has been used as a richer and more faithful way to study grounding, especially for abstract concepts \citep{barsalou2005situating,mcrae2005semantic}.

To measure how closely LLMs align with humans on conceptual grounding, we replicate two human property-generation studies on abstract concepts: \citet{harpaintner2018semantic} and \citet{kelly2024conceptual}. We selected these studies because both provide publicly available data and explicit coding taxonomies, enabling controlled human--model comparison under complementary experimental designs. \citet{harpaintner2018semantic} (\textit{Experiment 1}) organizes properties of abstract concepts into \textit{sensorimotor}, \textit{internal-state}, \textit{social}, and \textit{verbal-association} content, reflecting dimensions of abstract-concept representation that are well established in the broader literature \citep{troche2017,villani2019}. \citet{kelly2024conceptual} (\textit{Experiment 2}) provides a more specialized setting, with a different stimulus design and coding taxonomy with a particular focus on emotion concepts. Using both experiments allows us to test whether human--model grounding gaps persist across distinct property-generation paradigms rather than arising from a single dataset or coding scheme.


\subsection{Experimental setup}
\label{sec:exp-setup}

\begin{figure}[t]
  \centering
  \includegraphics[width=0.9\linewidth]{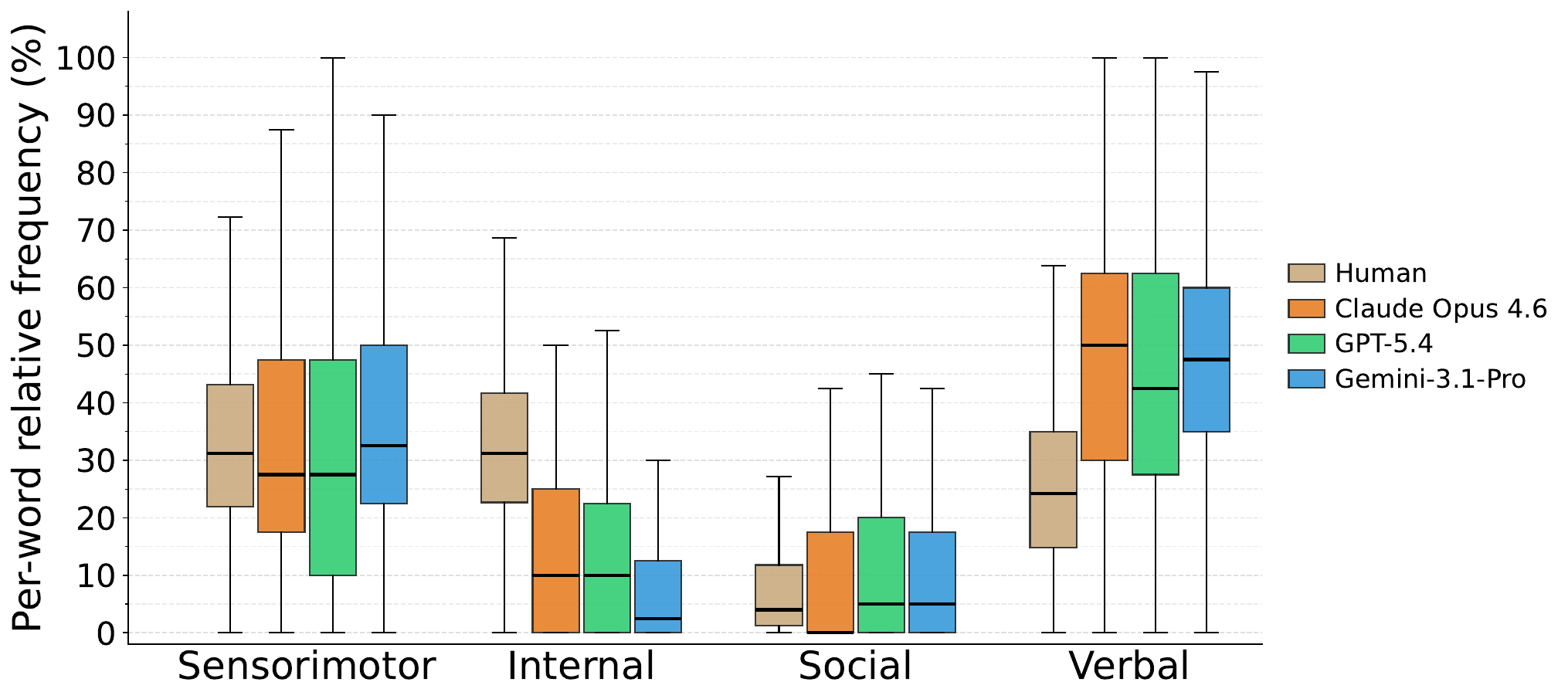}
  \caption{Per-word property-category frequency distributions for three frontier LLMs vs.\ the human baseline on the 293-abstract-noun benchmark \citep{harpaintner2018semantic}. Each box spans the 25th--75th percentile of word-level frequencies; median annotated.}
  \label{fig:main-boxplot}
\end{figure}

We apply the same procedure in both experiments, following their original designs. For each stimulus word, we prompt the model to generate the properties, situations, or associations that come to mind (four properties for Experiment~1 and five for Experiment~2), using the original human instructions (prompts in Appendix~\ref{app:prompts}). We then map the generated properties to the taxonomy of the corresponding human study and compare the resulting per-word category generations against the human ones. 
All scores are averaged over 10 runs per model; see Appendix~\ref{sec:appendix-variance} for convergence analysis.

\paragraph{Stimulus sets.}
The stimulus set consists of the target words presented to participants---or, in our case, to the models---for property generation. 
For the first experiment \citep{harpaintner2018semantic}, we derive a 293-word English stimulus set from the original 296 German abstract concepts. Two items were excluded as translation duplicates (the original paper provided English translations for all words), and one additional item was excluded because it is used as a one-shot example in the original prompt. In the second experiment \citep{kelly2024conceptual}, the stimulus set contains \fix{354} words split into three subsets: 118 abstract emotion words, \fix{117} abstract non-emotion words, and 119 concrete words. We use the \fix{235} abstract words for our analysis; see Appendix~\ref{app:exp2} for findings for concrete words.

%

\paragraph{Property coding.}
The generated properties are open-ended, so they must be assigned to the grounding categories defined by each study. Following the original coding schemes, generated properties for \citet{harpaintner2018semantic} are assigned to: \textit{Sensorimotor}, \textit{Internal State \& Emotion}, \textit{Social}, and \textit{Verbal Association}, while for \citet{kelly2024conceptual} to \textit{Taxonomic}, \textit{Entity}, \textit{Situation}, and \textit{Introspective}. 
We use LLMs-as-coders, after validating candidate models on human-annotated word-to-category pairs. The selected coders achieve agreement with the ground truth that is comparable to, or higher than, human coders, while remaining practical to use at scale. We use \textit{Gemini-2.5-Flash-Lite} as the primary coder for Experiment~1 and \textit{Gemini-2.5-Flash} for Experiment~2. Detailed coder comparisons, annotation protocols, and reliability analyses are reported in Appendix~\ref{app:coding}.

\paragraph{Alignment metric and human ceiling.}
After coding, each stimulus word is represented by the relative frequencies of its responses across the four target categories. We compare human and model frequency profiles using per-category Pearson~$r$, and report \textit{Mean~$r$}, the average across the four categories, as our main alignment metric. \rev{Concretely, for each category we correlate the human and model frequencies across all stimulus words (e.g., the Sensorimotor frequencies of the 293 concepts in Experiment~1) and then average the four per-category correlations. Our conclusions are unchanged under Spearman correlation and representational similarity analysis (Appendix~\ref{app:metrics}).} The corresponding human ceiling is the Pearson~$r$ that two independent splits of raters would obtain on the same words:
for Experiment~2 we compute it directly from \citet{kelly2024conceptual}'s participant responses, and for Experiment~1 we estimate it analytically from \citet{harpaintner2018semantic}'s per-word aggregates using standard inter-rater reliability formulas; the two methods agree to within \fix{$0.02$} on Experiment~2, validating our estimator (Appendix~\ref{app:ceilings}).

\paragraph{Models and settings.} 
We evaluated a diverse set of 21 frontier and open-weight LLMs spanning several families; we refer to Appendix~\ref{app:model-list} for the full list of models. We used the default temperature for all generations and to ensure coding reliability LLM-coders used a temperature of $T=0.0$.


\begin{table}[b]

  \centering
  \small
  \caption{Per-category Pearson correlations on Experiment 1 for three representative frontier models. 
    }
      \label{tab:harpaintner-main-models}
  \begin{tabular}{lccccc}
    \toprule
    Model & Mean $r$ & Sensorimotor & Internal & Social & Verbal \\
    \midrule
    \emph{Estimated human ceiling} & \emph{0.974} & \emph{0.966} & \emph{0.964} & \emph{0.971} & \emph{0.992} \\
    \midrule
    GPT-5.4                & 0.301 $\pm$ 0.005 & \fix{0.332} & \fix{0.258} & \fix{0.400} & \fix{0.213} \\
    Claude Opus 4.6        & 0.264 $\pm$ 0.004 & 0.203 & \fix{0.205} & \fix{0.362} & \fix{0.284} \\
    Gemini 3.1 Pro Preview & \fix{0.256} $\pm$ 0.008 & \fix{0.307} & \fix{0.163} & \fix{0.383} & \fix{0.171} \\

    \bottomrule
  \end{tabular}
  \par\smallskip
    
\end{table}

\subsection{Results}
\label{sec:behavioral-results}
\paragraph{Experiment 1.}
\label{sec:exp1}

\begin{figure}[t]
  \centering
  \includegraphics[width=0.9\linewidth]{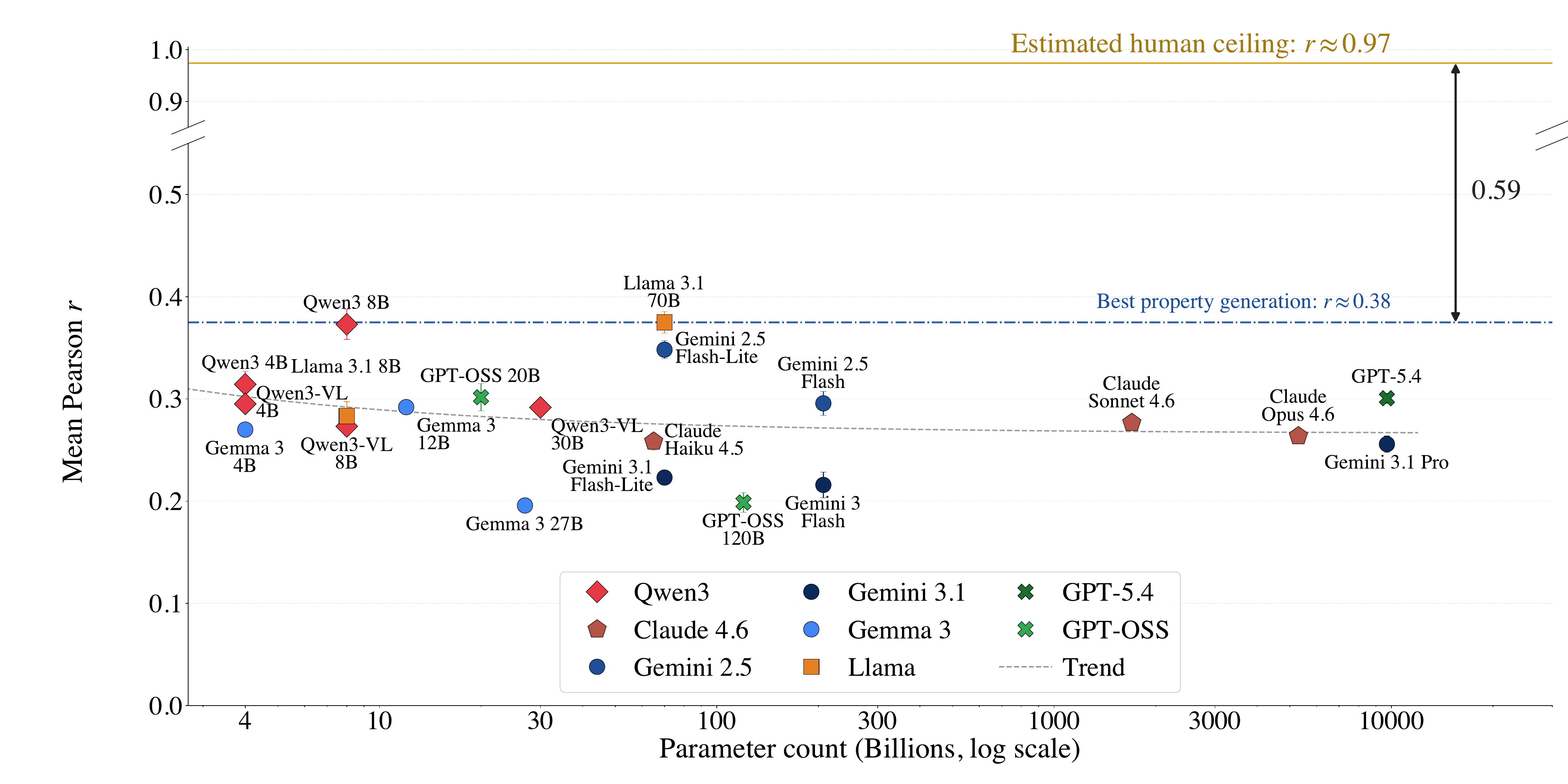}
  \caption{Mean r for all LLMs vs.\ the human ceiling on Experiment 1 \citep{harpaintner2018semantic}. Parameter count for closed models is based on estimations from \citet{li2026incompressible}. \fix{The dashed curve shows the saturating fit to the 21 models (trend). The dash-dotted line marks the best model, and the arrow its distance to the human ceiling.}}
  \label{fig:harpaintner-main}
\end{figure}

Figure~\ref{fig:main-boxplot} illustrates the per-word and per-category distributions for human responses and for three representative frontier models: Claude Opus 4.6, Gemini 3.1 Pro Preview, and GPT-5.4. Relative to humans, all three models generate substantially more \textit{Verbal Association} properties and substantially fewer \textit{Internal State \& Emotion} properties.  The same overall pattern holds across all models (see Appendix~\ref{app:exp1} for the full set of models). The similarity between the frequency of the different models also indicates that this is not a coincidental property of a single model family, but a consistent behavior across current architectures. By contrast, \textit{Sensorimotor} frequencies remain closer to the human range in the aggregate, and \textit{Social} frequencies are also broadly comparable, although this category is relatively sparse for both humans and models and is therefore less informative at the frequency level alone. 

The frequency distributions indicate a systematic mismatch between human and model grounding, but aggregate frequency alone does not show whether models ground the \emph{same} concepts in the \emph{same} way as humans. For example, a model may produce a human-like number of \textit{Sensorimotor} or \textit{Social} properties overall, while assigning those properties to different words. We therefore measure alignment directly by computing Pearson correlations between human and model category-frequency generations across the stimulus set, as described in Section~\ref{sec:exp-setup}.

As shown in Figure~\ref{fig:harpaintner-main}, human-model alignment remains far below the estimated human--human ceiling across all 21 models.
The best model reaches only a mean \fix{$r \approx 0.38$}, compared with an estimated human ceiling of about $0.97$, with the same pattern across individual categories as shown in Table~\ref{tab:harpaintner-main-models}. Thus, even when aggregate frequencies appear relatively human-like, models often associate grounding dimensions with different concepts than humans do. This gap is not explained by scale: larger models and stronger general-purpose systems do not show reliably higher human alignment, suggesting that the mismatch is not simply a capability bottleneck that shrinks with model size.

\begin{figure}[t]
    \centering
    \begin{subfigure}[t]{0.45\textwidth}
        \centering
        \includegraphics[width=\linewidth]{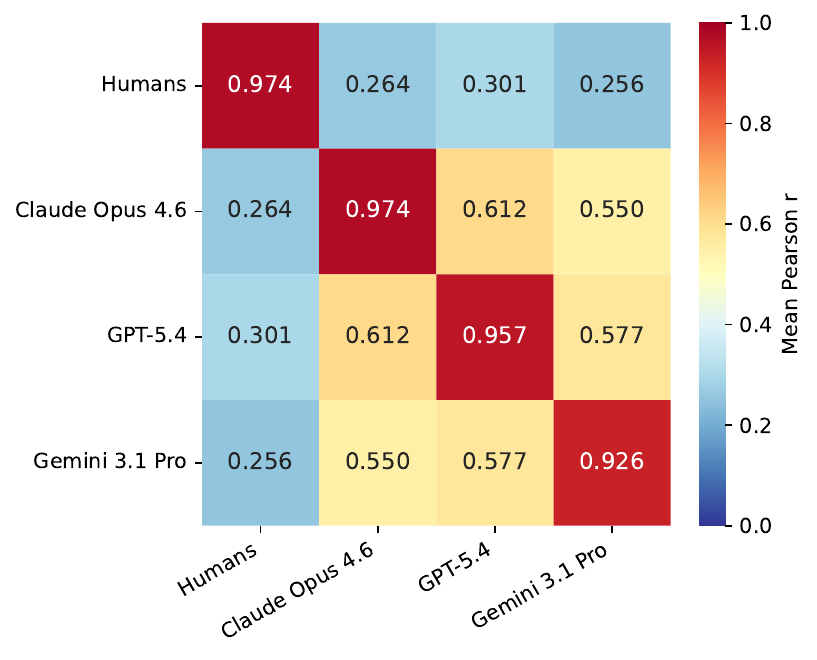}
        \caption{Experiment 1 (\citet{harpaintner2018semantic})}
         
        \label{fig:inter_model_heatmap_harpaintner}
    \end{subfigure}
    \hfill
    \begin{subfigure}[t]{0.45\textwidth}
        \centering
        \includegraphics[width=\linewidth]{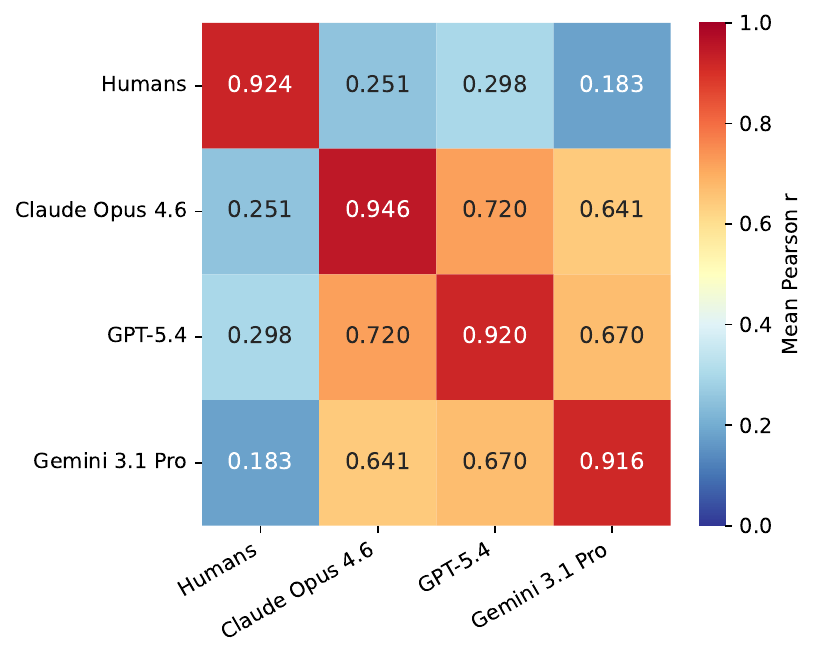}
        \caption{Experiment 2 (\citet{kelly2024conceptual})}
        \label{fig:inter_model_heatmap_kelly}
    \end{subfigure}
    \caption{Pearson-$r$ correlation heatmaps for three representative frontier models and humans.}
    \label{fig:inter_model_heatmaps}
\end{figure}

Inter-model correlations reveal a complementary pattern. As shown in Figure~\ref{fig:inter_model_heatmap_harpaintner}, models correlate substantially more with one another than with humans, despite spanning different providers, data mixtures, and architectural choices. This suggests that current LLMs share a rather common mode of abstract-concept grounding, but one that is systematically distinct from the human pattern. The grounding gap is therefore not idiosyncratic to a single model family, but shared across \rev{the instruction-tuned LLMs we evaluated. This shared pattern may partly reflect overlapping pretraining corpora and post-training pipelines rather than a property of language modeling as such.}


\paragraph{Experiment 2.}
\label{sec:exp2}
Experiment~2 confirms the findings of Experiment~1 under a different experimental setup. Human-model alignment remains low in the combined abstract condition, which pools the \textit{abstract emotion} and \textit{abstract non-emotion} items: even the strongest models reach only a mean \fix{$r \approx 0.37$}, far below the human-to-human correlation of \fix{$r \approx 0.92$}. Full results for all models are reported in \fix{Appendix~\ref{app:exp2}}.

Inter-model correlations also follow the same pattern as in Experiment~1. As shown in Figure~\ref{fig:inter_model_heatmap_kelly}, models correlate substantially more with one another than with humans. Together, the two experiments reveal a robust behavioral grounding gap: current LLMs share a common model-like pattern of abstract-concept representation, but one that remains systematically distinct from humans.

\subsection{LLM-coder error margins}
\label{sec:llm-coder}

The \fix{Experiment~1} Pearson correlations and category frequencies reported above are based on labels produced by Gemini~$2.5$~Flash-Lite. To estimate whether this affects our conclusions, we manually re-coded the outputs of  three frontier models \rev{(two of the authors served as annotators, coding each property independently; agreement $68.9\%$, Cohen's $\kappa=0.59$; Appendix~\ref{app:coding-bench})} and compared the resulting metrics with those obtained from Flash-Lite labels on the same property pairs. The differences are minor: replacing Flash-Lite with human coders changes Mean~$r$ by only $\Delta = +0.014 \pm 0.019$ on average, with no consistent per-model shift (per-model $95\%$ confidence intervals contain zero), and changes per-category mean frequencies by roughly $\pm 2\%$. These margins are far smaller than the human--model gap reported in Section~\ref{sec:exp1}, indicating that our main results are not driven by the use of LLMs as coders. Detailed analyses are reported in Appendix~\ref{app:coding}.

\rev{Appendices~\ref{app:budget} to~\ref{app:multimodal} further examine the property budget, prompt framing, instruction tuning, statistical units and multimodality; none of these closes the gap.}

\section{Rating analysis of grounding dimensions}
\label{sec:rating-exp}

To better understand the grounding gap identified through the previous experiments, we study whether this gap arises because current LLMs do not properly recover the grounding dimensions that structure human concept representations, or because these dimensions are not recruited in a human-like way when concepts are generated freely. To distinguish between these possibilities, we replicate a rating experiment from cognitive science \citep{troche2017}. Unlike property generation, rating experiments do not test the spontaneous associations for a concept; instead, they test whether the model can score a word on semantic dimensions once these dimensions are explicitly specified. 

\paragraph{Setup.}
We follow the setup of \citet{troche2017}, in which human participants rated 751 English nouns on 14 dimensions using a 1--7 Likert scale. The dimensions cover sensory content (\textit{Color}, \textit{Taste/Smell}, \textit{Tactile}, \textit{Visual Form}, \textit{Auditory}), motor content (\textit{Self-Motion}), internal and evaluative content (\textit{Emotion}, \textit{Polarity}, \textit{Morality}, \textit{Thought}), social content (\textit{Social}), and magnitude-related content (\textit{Space}, \textit{Quantity}, \textit{Time}).
Using the original prompt "I relate this word to [\emph{X}]," we evaluated the same 21 models from previous experiments over 10 shuffled runs each. Performance is reported as Mean~r, the Pearson correlation between model and human ratings averaged across all 14 dimensions.

\begin{figure}[t]
  \centering
  \includegraphics[width=0.9\linewidth]{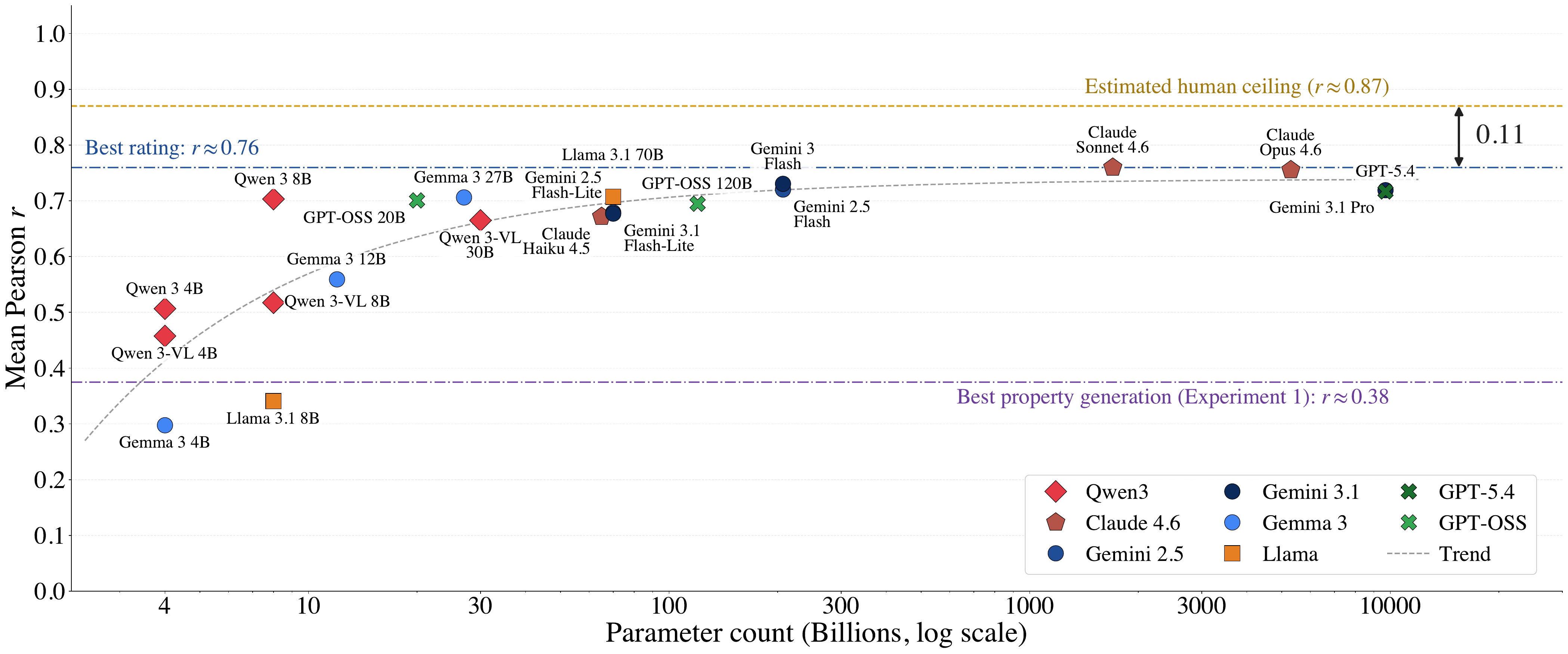}
  \caption{Mean r for all LLMs vs.\ the human ceiling on the rating experiment \citep{troche2017}. Parameter count for closed models is based on estimations from \citet{li2026incompressible}. \fix{The dashed curve shows the saturating fit to the 21 models (trend). The dash-dotted lines mark the best model on this task (blue) and on Experiment~1 (purple), and the arrow the distance from the best model to the human ceiling.}}
  \label{fig:troche-scaling-plot}
\end{figure}

\paragraph{Results.}
The results differ sharply from those of property generation (compare Figures~\ref{fig:harpaintner-main} and \ref{fig:troche-scaling-plot}). On this task, recent frontier LLMs align much more closely with human judgments: the strongest model reaches Mean~$r \approx 0.76$ against an estimated human ceiling of about \fix{$0.87$}, and alignment improves substantially with model scale. \rev{This contrast holds when controlling for the word pool, the number of dimensions and the human ceilings: on the 93 words shared by both tasks, models reach $78.4\%$ of the human ceiling in rating but only $31.8\%$ in property generation (Appendix~\ref{app:matched}).} It suggests that the grounding gap is not due to an inability to recognize dimensions such as sensation, emotion, or sociality: current LLMs can recover these dimensions when they are made explicit in the task, but do not recruit them in a human-like way when asked to generate the content of abstract concepts freely. Extended experimental details and results are provided in Appendix~\ref{app:exp-ratings}.


\section{Mechanistic analysis}
\label{sec:sae}

Having established that LLMs can predict grounding dimension ratings, we investigate whether they internally encode features tracking these dimensions by analyzing the 4B and 12B Gemma 3 instruction-tuned models using Gemma Scope 2 sparse autoencoders (SAEs)~\citep{gemmascope2_2025}. We isolate latent features correlated with grounding dimensions, focusing on those that align strongly with human experiential categories. \rev{For some dimensions, the best-matching feature correlates with the human ratings more strongly than the same model's explicit ratings do. Because the analysis covers two models from a single family, we treat these results as exploratory.} While steering these representations predictably increases the generation of target-category properties, we view this merely as a methodological sanity check; artificially amplifying feature activations does not close the grounding gap, as it overrides rather than reflects the model's natural internal structure.




\subsection{Methodology}
\label{sec:sae-methods}


\paragraph{Feature identification.}
\label{sec:feature-identify}
We ask whether the model contains SAE features that correspond to grounding categories or sub-groups beyond individual words. First, we collect a pool of \fix{425} labeled nouns from four psycholinguistic norm databases \citep{lynott2020lancaster, diveica2023quantifying, scott2019glasgow, brysbaert2014concreteness}, each associated with one of the 4 grounding dimensions of interest (sensorimotor, internal-state, social, and abstract content). 
For each noun $w$, we generate 10 English sentences ending in $w$ using Gemini-3.1-pro-preview, and then remove $w$, creating a form of cloze (fill-the-blank) questions to study model activations when generating $w$. We record the SAE activation at the final token of these cloze questions, i.e.\ the position where the model is about to generate $w$, and take the per-feature median across the ten sentences as the noun's activation signature. A feature enters category $c$'s candidate pool if it has non-zero median activation on at least five distinct nouns from $c$. The approach is similar to the one used by~\citep{sofroniew2026emotions} for emotion vectors. Construction details are reported in Appendix~\ref{app:feat-identification}. This approach gives us a set of candidate features for all 4 grounding dimensions. 

\paragraph{Tracking grounding dimension norms with SAE features.}
The previous stage is designed to select features that their activations are correlated with generation of nouns related to their category, but their activations are not necessarily aligned with humans. Our goal is to find features that not only respond to a grounding dimension, but do so in a way that is aligned with human norms in a grounding dimension.
To achieve this we evaluate these candidate features on the same dataset we used in our rating experiment in Section~\ref{sec:rating-exp}. For the \emph{feature} measurement, the original rating prompts are unsuitable: they target a specific dimension and elicit a numeric answer, whereas our candidate features were identified in the regime of free noun generation. We instead use two simple and generic property-generation prompts and the original complex property-generation prompt from Section~\ref{sec:experiments} and average the results, record each candidate feature's activation at the final prompt position, and correlate it with the human ratings on each dimension. This measures how well the feature itself tracks the human grounding signal while simulating the property-generation task.

\begin{figure}[t]
  \centering
  \includegraphics[width=0.92\linewidth]{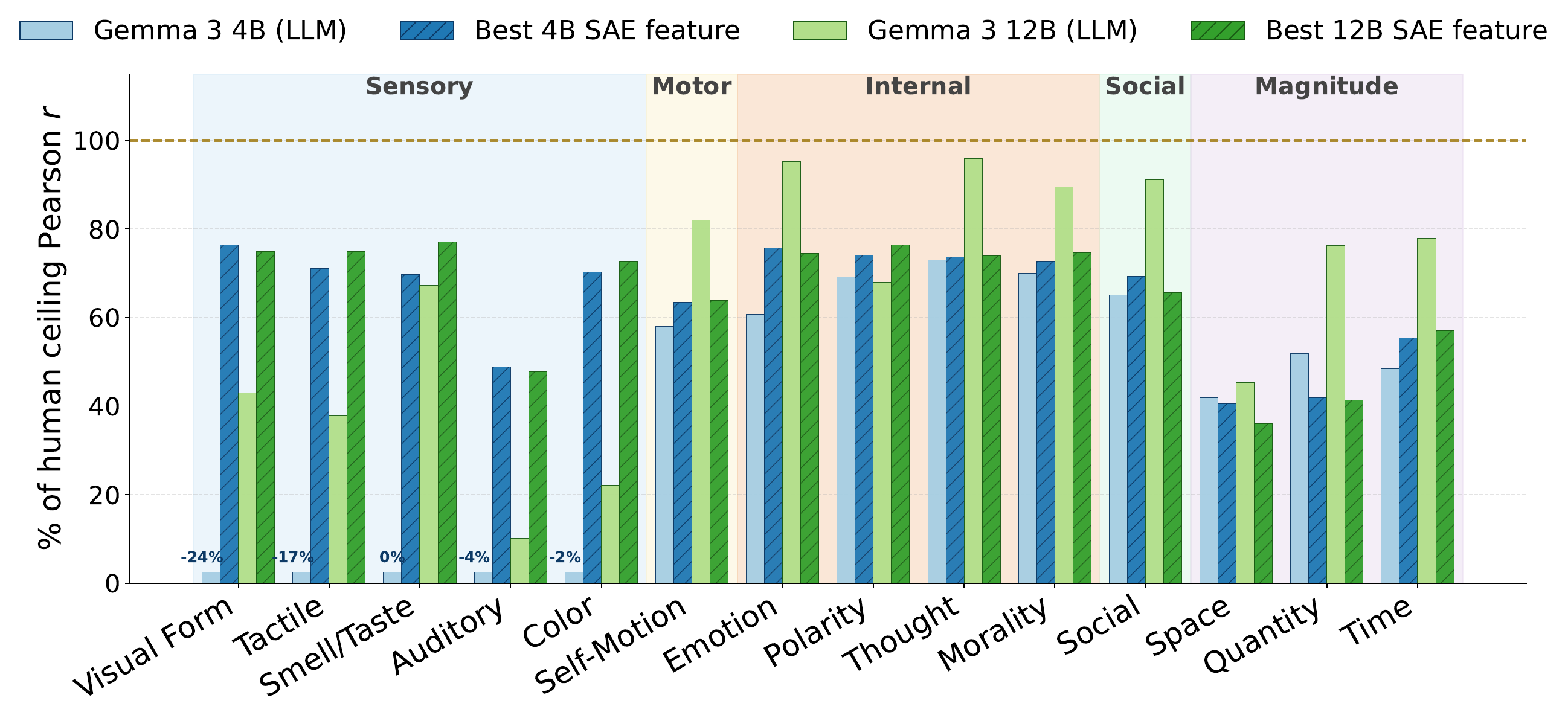}
  \caption{Percentage of mean r for all LLMs and SAE features vs.\ the estimated human ceiling mean r on the rating experiment dimensions.}
  \label{fig:sae}
\end{figure}

\subsection{Results}
\label{sec:sae-results}
Overall, both models show a rather strong behavioral alignment with human ratings across most grounding dimensions (Figure~\ref{fig:sae}). Sensory dimensions are the main exception: model ratings remain far below the human ceiling, which is consistent with the findings of \citet{xu2025large}. Nevertheless, the SAE analysis identifies features that correlate with all evaluated dimensions, including the behaviorally weaker sensory ones. For most dimensions, the best-matching SAE features reach Pearson correlations in the range of $r=0.6$ to $0.8$, suggesting that these grounding-related dimensions are at least partially reflected in the models' internal representations.
The main exception is the magnitude category, which includes space, quantity, and time and shows weaker feature alignment, possibly reflecting the more abstract nature of these dimensions. Overall, these results suggest that the models contain internal features that track several sub-dimensions of grounding, with these features being aligned with human judgments to some extent, and should not be interpreted as complete or fully human-like representations.


\subsection{Feature steering}
\label{sec:sae-results-steer}

To interpret and validate these features, we conduct a steering experiment on Gemma~3~4B using the highest-correlation SAE feature for each dimension: sensory, motor, internal, social, and magnitude. We replicate Experiment~1 (Section~\ref{sec:exp1}) under feature steering, using feature clamping \citep{templeton2024scaling} to a fixed high activation value. Steering increases target-category property generation in all cases: sensory (+5.3\% \emph{Sensorimotor}), motor (+4.7\% \emph{Sensorimotor}), internal (+9.4\% \emph{Internal}), social (+14.7\% \emph{Social}), and magnitude (+1.7\% \emph{Verbal}). These effects support the feature--category associations identified in the correlation analysis.
We further inspect the highest-activating examples for each feature using Neuronpedia,\footnote{Neuronpedia is an open-source platform for SAE interpretability and visualization: \url{https://www.neuronpedia.org}.} and find that the features generally activate on words related to their grounding dimensions. However, they are not perfectly category-specific: some also activate on words outside the narrow target dimension, suggesting that they capture broader semantic patterns rather than clean, isolated grounding categories or sub-dimensions. For details, see Appendix~\ref{app:feat-steer}.

\section{Discussion}
\label{sec:discussion}

Grounding is important for humans because it connects language to lived experience and the physical world. As LLMs take roles originally meant for humans and move closer to agentic autonomy, 
their apparent differences from humans in grounding acquire some significance. The grounding gap we identified \rev{may not} go away through scaling and text-based training alone, since (a) \rev{among the instruction-tuned models we tested,} scaling \citep{NEURIPS2022_c1e2faff, emergent_2022} does not appear to help, with state-of-the-art models remaining remarkably far from human ceilings, and (b) the excessive reliance on verbal associations we observe is \rev{consistent with} training that is focused on prediction from linguistic context. \rev{Our base-model analysis (Appendix~\ref{app:base}) shows that post-training also contributes: pretrained-only checkpoints are moderately more aligned, although they retain the same qualitative profile. We cannot cleanly separate the contributions of pretraining and post-training, and our evidence directly supports a grounding gap in current post-trained models.}
Genuine progress on this front may require fundamental shifts, such as novel training techniques or alternative architectures \citep{fung2025embodied, brooks1991intelligence}.

\rev{In the two Gemma 3 models we analyzed, some SAE features track grounding dimensions, in some cases more closely than the model's own explicit ratings.} However, these features do not exclusively encode sensory, internal, or social experiences (see Appendix~\ref{app:feat-steer} for details). According to one mainstream theory of grounded cognition \citep{barsalou1999perceptual}, true conceptual understanding requires the capacity to internally simulate physical experiences. Whether LLMs possess, or can acquire through training, internal circuits capable of such simulation remains an open challenge for future research.

\paragraph{Limitations.}
Our behavioral analysis is limited by the generality of the property-generation experiments we replicate, although both are well established in the literature. 
To our knowledge, these are the only extant property-generation studies on abstract-concept grounding with both openly available stimulus sets and coding taxonomies suitable for direct replication and comparison.  
The fact that we observe highly consistent results across these two independently designed experiments, with different stimuli, taxonomies, and participant samples, suggests that the observed gap is not specific to a single experimental setup. 

A second limitation is the use of LLMs as coders, which could introduce model-specific biases into the estimated category frequencies and correlations. We mitigate this by validating candidate coders against human-annotated word-to-category pairs and selecting models with human-level or stronger agreement. We also show in Section~\ref{sec:llm-coder} that replacing the LLM coder with human coders changes the main metrics only marginally, with differences far smaller than the human--model gaps reported above. This makes it highly unlikely that our conclusions are driven by the coding procedure.

\rev{Third, our base-versus-instruct comparison is not a clean causal ablation of post-training\fix{, because base models need few-shot demonstrations to follow the task, and these demonstrations can themselves shift the category distribution} (Appendix~\ref{app:base}).} \fix{The post-training recipe is also undisclosed for most models we study, and varies widely across those that document it, so we cannot attribute behavioral shifts to specific post-training steps.}

Finally, our SAE analysis is limited to Gemma models due to the limited availability of high-quality SAEs for recent LLMs and the cost of scaling the analysis. \rev{Both models belong to the same family and share data, architecture, and training pipeline, so the features we find may reflect family-specific structure rather than a pattern that generalizes across LLMs.} More generally, SAE features are interpretable approximations of model representations, not complete decompositions. The absence of a feature in an SAE therefore does not guarantee that the corresponding information is absent from the underlying model. We therefore treat the SAE results as exploratory evidence that complements the behavioral findings, rather than as the basis for strong mechanistic claims.

\section{Conclusion}
\label{sec:conclusion}

We investigate how well frontier LLMs are cognitively aligned with humans in their grounding of abstract concepts. We replicate two property-generation experiments on 21 LLMs, and find strong evidence that \textbf{LLMs are not grounded in the same way humans are} --- but of course they were not trained to be. In particular, LLMs exhibit systematic under-production of internal-state properties and over-production of \rev{verbal associations}, and overall very poor Pearson correlation with humans, while they correlate well with each other. These failures are uniform across \rev{the instruction-tuned} LLM families \rev{we tested} and do not improve with scale. 
In a replicated rating experiment, LLMs exhibit a reasonable understanding of concept categories, and \rev{in Gemma-3-4B and Gemma-3-12B we find SAE features whose activations track these categories, suggesting that grounding-related information can be present internally even when it is not expressed in free generation.} 
Going forward, our results suggest that grounding alignment with humans may be an interesting and novel quantity that is invisible to current testing benchmarks, and may be worth watching in the future.  
And finally, \rev{our results suggest that} further scaling or better text-based training \rev{alone may} not close the gap, so we may need to rethink the way we design and train LLMs.

\begin{ack}
\rev{O.~S. Chlapanis was supported by project MIS 5154714 of the National Recovery and Resilience Plan Greece 2.0, funded by the European Union under NextGenerationEU. O. Menis Mastromichalakis was supported by the project DECOLLAGE (ERC-2022-CoG 101088763) and through national funds and co-funded EU funds under UID/50008: Instituto de Telecomunica\c{c}\~oes. C.~H. Papadimitriou was supported by NSF grants CCF-2212233 and CCF-2332922, by ONR MURI grant N00014-25-1-2116, by the NSF AI Institute for Artificial and Natural Intelligence, and by the ERC grant NEAGT (101267783).}
\end{ack}

\bibliographystyle{abbrvnat}
\bibliography{references}


\newpage

\appendix

\section{LLMs-as-coders for property-generation experiments}
\label{app:coding}

\subsection{Benchmarking LLMs-as-coders}
\label{app:coding-bench}

In this section, we evaluate the performance of LLMs on the downstream task of automated property coding. Our methodology builds upon the established ``LLM-as-a-judge'' paradigm \citep{zheng2023judging}. However, deploying LLMs as coders presents a more straightforward, and arguably more reliable, framework than using them as evaluative judges. A typical LLM-judge must typically surpass the candidate model in reasoning capabilities to accurately score its output, while a LLM-coder merely has to be able to classify the properties properly. This coding task is fundamentally a simplified variation of the rating task detailed in Section \ref{sec:rating-exp}, a domain where we have already demonstrated robust LLM performance. Furthermore, because this task involves classification rather than quality evaluation, it avoids the self-enhancement bias typically observed in LLM-as-judge setups \citep{zheng2023judging}. This lack of self-bias makes it methodologically sound to use the same model to both generate and subsequently code its own responses.

\paragraph{Data collection and benchmark creation.}
To rigorously benchmark the models as coders, we needed a substantial dataset of coded properties. Fortunately, \citet{kelly2024conceptual} provides a large-scale public dataset: the complete $49{,}942$ word-property pairs produced in their experiment. For experiment~1, no such word-property pairs were provided, so \fix{two of the authors independently labeled} 2,077 pairs \rev{(agreement $68.9\%$, Cohen's $\kappa = 0.59$)}, which were generated by the three frontier LLMs (Claude-Opus 4.6, \fix{GPT-5.4}, \fix{Gemini-3.1-Pro}). We utilized these two datasets as our primary benchmarks to identify the optimal LLM coder for each experiment.

\paragraph{Validation metrics and human reference.}
For each candidate coder we report percentage agreement with the ground-truth label and Cohen's $\kappa$. The human reference for Experiment~1 is the $76.8\%$ two-coder joint agreement reported by \citet{harpaintner2018semantic} on their original validation subset\fix{, and our two annotators agreed on $68.9\%$}; \citet{harpaintner2018semantic} do not report a $\kappa$ ceiling, so we leave that cell blank. For Experiment~2, the human reference is the $64.25\%$ two-coder agreement and $\kappa = 0.52$ reported by \citet{kelly2024conceptual}.

To ensure deterministic coding results, we use temperature $T=0.0$ for all LLM-coders. 

\begin{table}[ht]
\centering
\small
\caption{Experiment~1 coder leaderboard. Six LLM coders predict the 4-category property label on the 2077-pair ground truth. \% Agreement is overall percentage agreement, $\kappa$ is Cohen's kappa. Per-category columns are recall, i.e.\ the percentage of pairs labeled by the human as that category that the LLM also labeled as that category. $^{\dagger}$Human ceiling is the 76.8\% two-coder joint agreement reported by \citet{harpaintner2018semantic}. \textbf{Gemini-2.5-Flash-Lite} was selected as the coder for all experiments with this dataset, unless otherwise specified.}
\label{tab:exp1-leaderboard-coding}
\begin{tabular}{l c c cccc}
\toprule
Coder & \% Agreement & $\kappa$ & Sensorimotor & Emotion & Social & Association \\
\midrule
\emph{Human ceiling$^{\dagger}$} & \emph{76.8} & \emph{--} & \emph{--} & \emph{--} & \emph{--} & \emph{--} \\
\midrule
Gemini 3.1 Pro & 68.6 & 0.510 & 69.6 & 62.6 & 55.5 & 72.1 \\
Claude Sonnet 4.6 & 67.6 & 0.492 & 55.6 & 68.5 & 57.3 & 76.9 \\
\textbf{Gemini 2.5 Flash-Lite} & 67.2 & 0.505 & 66.2 & 75.9 & 62.1 & 67.1 \\
Gemini 2.5 Flash & 65.6 & 0.484 & 77.0 & 60.6 & 63.0 & 60.3 \\
GPT-OSS 120B & 64.4 & 0.477 & 76.0 & 66.5 & 68.6 & 55.9 \\
GPT-5 mini & 64.2 & 0.486 & 87.5 & 68.0 & 67.8 & 48.6 \\
\bottomrule
\end{tabular}

\end{table}


\begin{table}[t]
\centering
\small
\caption{Experiment 2 coder leaderboard. \% Agreement and $\kappa$ are computed against the \fix{$47{,}123$} gold human-coded labels. $^{\dagger}$Human ceiling is the 64.25\% two-coder agreement and $\kappa = 0.52$ reported by \citet{kelly2024conceptual}. \textbf{Gemini-2.5-Flash} was selected as the coder for all experiments with this dataset, unless otherwise specified.}
\label{tab:exp2-leaderboard-coding}
\begin{tabular}{l c c}
\toprule
Coder & \% Agreement & $\kappa$ \\
\midrule
\emph{Human ceiling$^{\dagger}$} & \emph{64.25} & \emph{0.52} \\
\midrule
\textbf{Gemini 2.5 Flash }& \fix{69.2} & \fix{0.577} \\
GPT-OSS 120B & \fix{64.1} & \fix{0.516} \\
Gemma 3 12B & 60.9 & 0.476 \\
Gemini 2.5 Flash-Lite & \fix{60.0} & \fix{0.482} \\
\bottomrule
\end{tabular}
\smallskip

\end{table}

\paragraph{Error margins for Gemini-2.5-Flash-Lite as a coder.}
\label{app:coding-error-margins}

The Experiment~1 coder benchmark in Table~\ref{tab:exp1-leaderboard-coding} measures property-level agreement, but the Experiment~1 leaderboard actually consumes the per-word category-frequency vector and its Pearson~$r$ against the human norms. To estimate the error coder substitution introduces at that leaderboard level, we re-coded the single hand-coded run of three reference generation models (Claude Opus~$4.6$, Gemini~$3.1$~Pro, GPT-$5.4$) with both Flash-Lite and the human ground truth on the exact same (word, property) pairs and bootstrapped the difference over words ($1000$ paired iterations, same resampled words for both coders within an iteration so that $\Delta = (\text{Flash-Lite} - \text{Human})$ is a paired statistic).
The Mean~$r$ a human coder would have produced has a cross-model average difference from the LLM-coder values of $\Delta = +0.014 \pm 0.019$, with every model $95\%$ CI containing zero.
On the same property pairs the cross-model average frequency $\Delta$'s are $+2.3$\% Sensorimotor, $-0.1$\% Internal, $-0.5$\% Social, $-1.7$\% Verbal Association, with every per-model $95\%$ CI containing zero. So the per-category attribution rates a human coder would have produced are within roughly $\pm 2$ percentage points of the Flash-Lite values.

\section{Extended results for Experiment 1}
\label{app:exp1}

Here we present extended quantitative results for the first experiment (Section~\ref{sec:exp1}) based on the \citet{harpaintner2018semantic} dataset. 

\subsection{Detailed model setup}
\label{app:model-list}
We evaluate $21$ LLMs spanning closed-weight frontier and open-weight families. All generations use each provider's default sampling temperature. Coding is performed at $T=0.0$ for determinism (Section~\ref{app:coding-bench}).

The closed-weight models are accessed through their official APIs: \textbf{Anthropic} (Claude Opus 4.6, Claude Sonnet 4.6, Claude Haiku 4.5), \textbf{OpenAI} (GPT-5.4), and \textbf{Google} (Gemini 3.1 Pro, Gemini 3 Flash, Gemini 3.1 Flash-Lite, Gemini 2.5 Flash, Gemini 2.5 Flash-Lite). The open-weight models are run on 8x Quadro RTX 6000 GPUs, with model-parallel sharding for the $70$B and $120$B systems: \textbf{Meta} (Llama 3.1 70B Instruct, Llama 3.1 8B Instruct), \textbf{OpenAI open-weights} (GPT-OSS 120B, GPT-OSS 20B), \textbf{Google open-weights} (Gemma-3-27B-IT, Gemma-3-12B-IT, Gemma-3-4B-IT), and \textbf{Alibaba Qwen} (Qwen3 8B, Qwen3 4B, Qwen3-VL 30B, Qwen3-VL 8B, Qwen3-VL 4B). 

Each model produces $10$ independent runs over the full $293$-word Harpaintner stimulus set. Per-word category-frequency vectors are averaged across runs before correlating with the human norms; $\pm$ values reported in the leaderboard are bootstrap standard deviations over $1000$ resamples of the runs (Section~\ref{sec:appendix-variance-bootstrap}).

\subsection{Detailed results for all 21 models}
\label{app:exp1-full-leaderboard}

\begin{figure}[p] 
    \centering
    
    \begin{minipage}{\textwidth}
        \centering
        
        \includegraphics[width=0.8\textwidth]{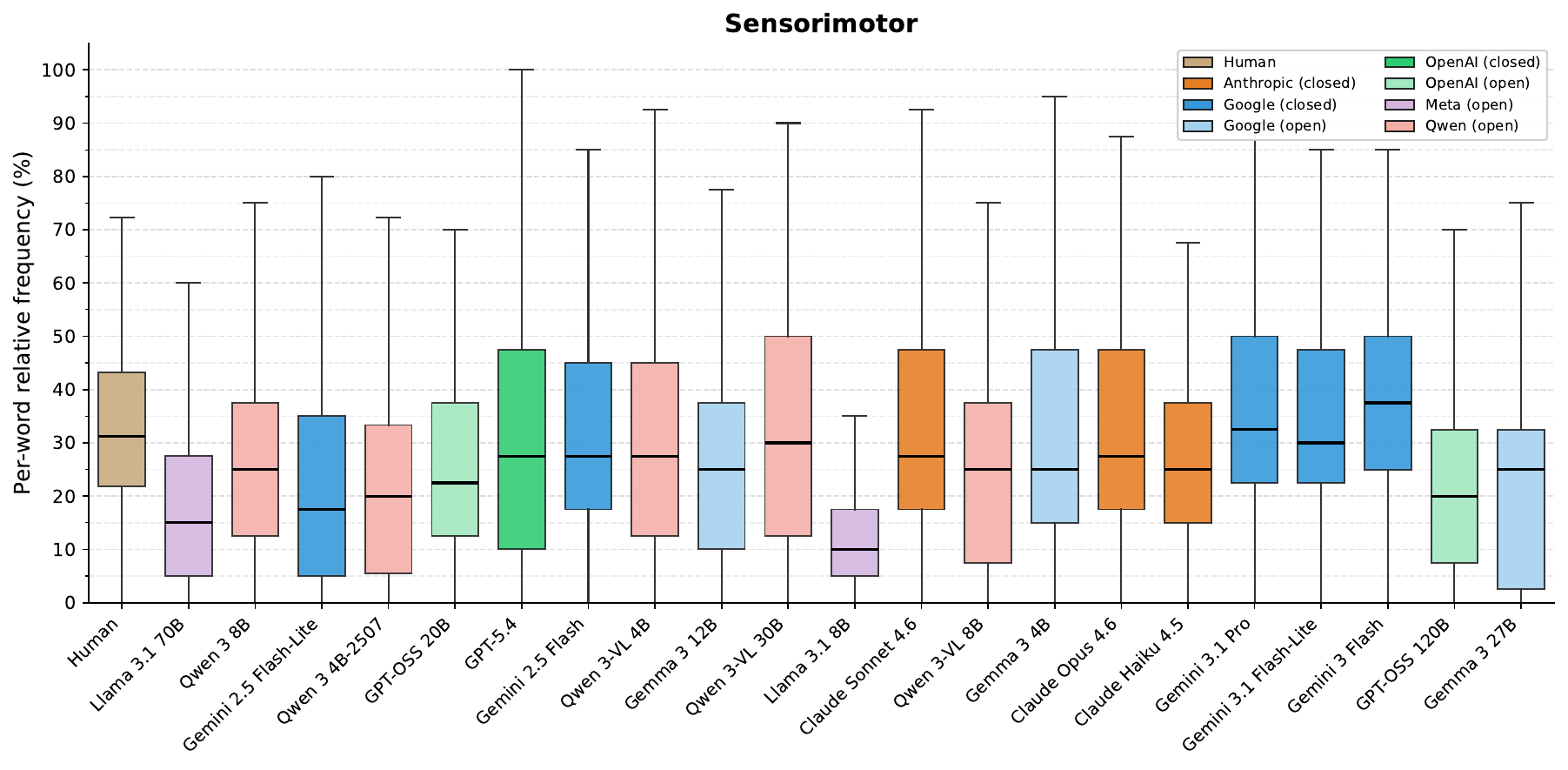}
        \vfill 
        
        \includegraphics[width=0.8\textwidth]{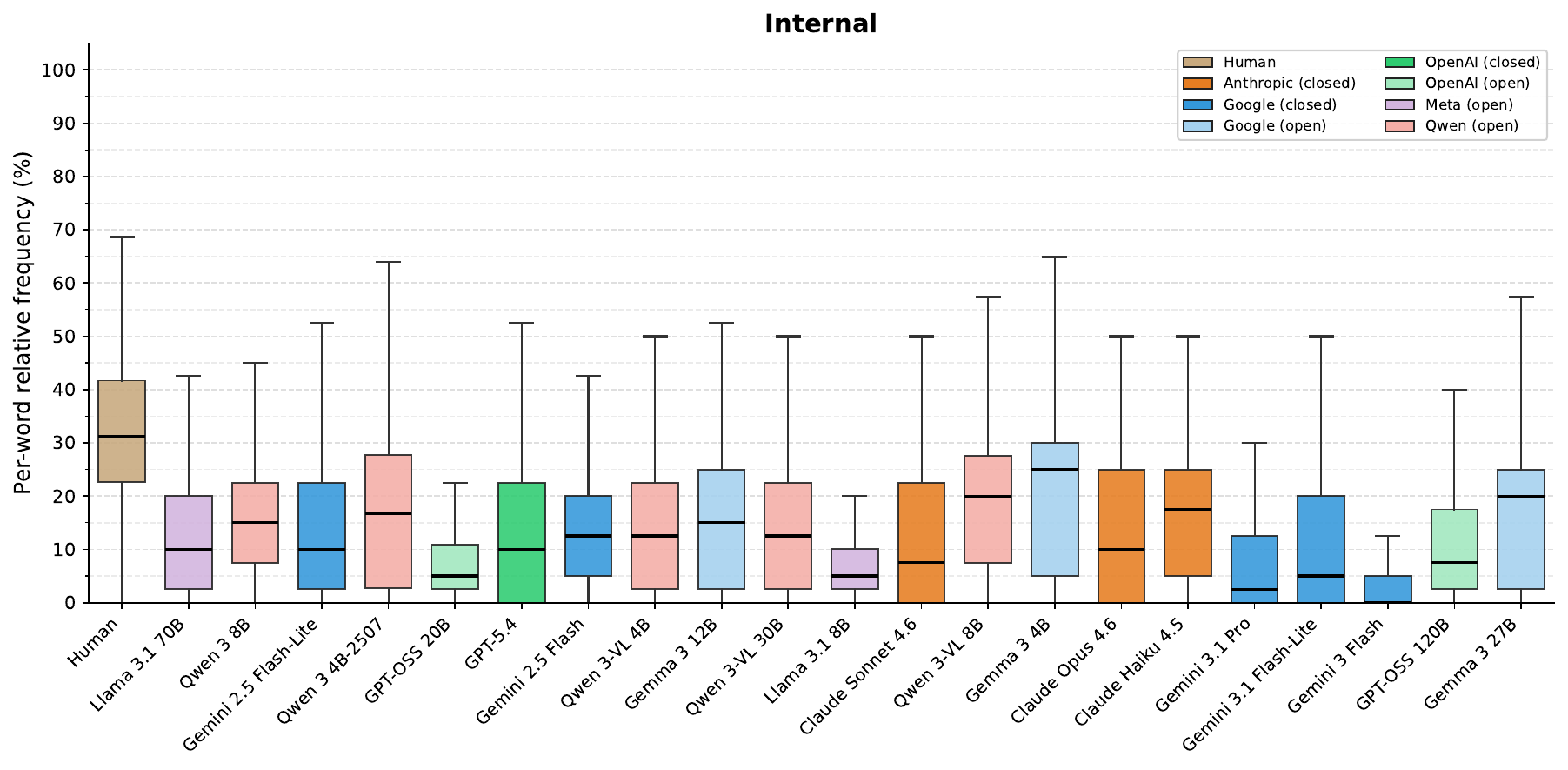}
        \vfill
        
        \includegraphics[width=0.8\textwidth]{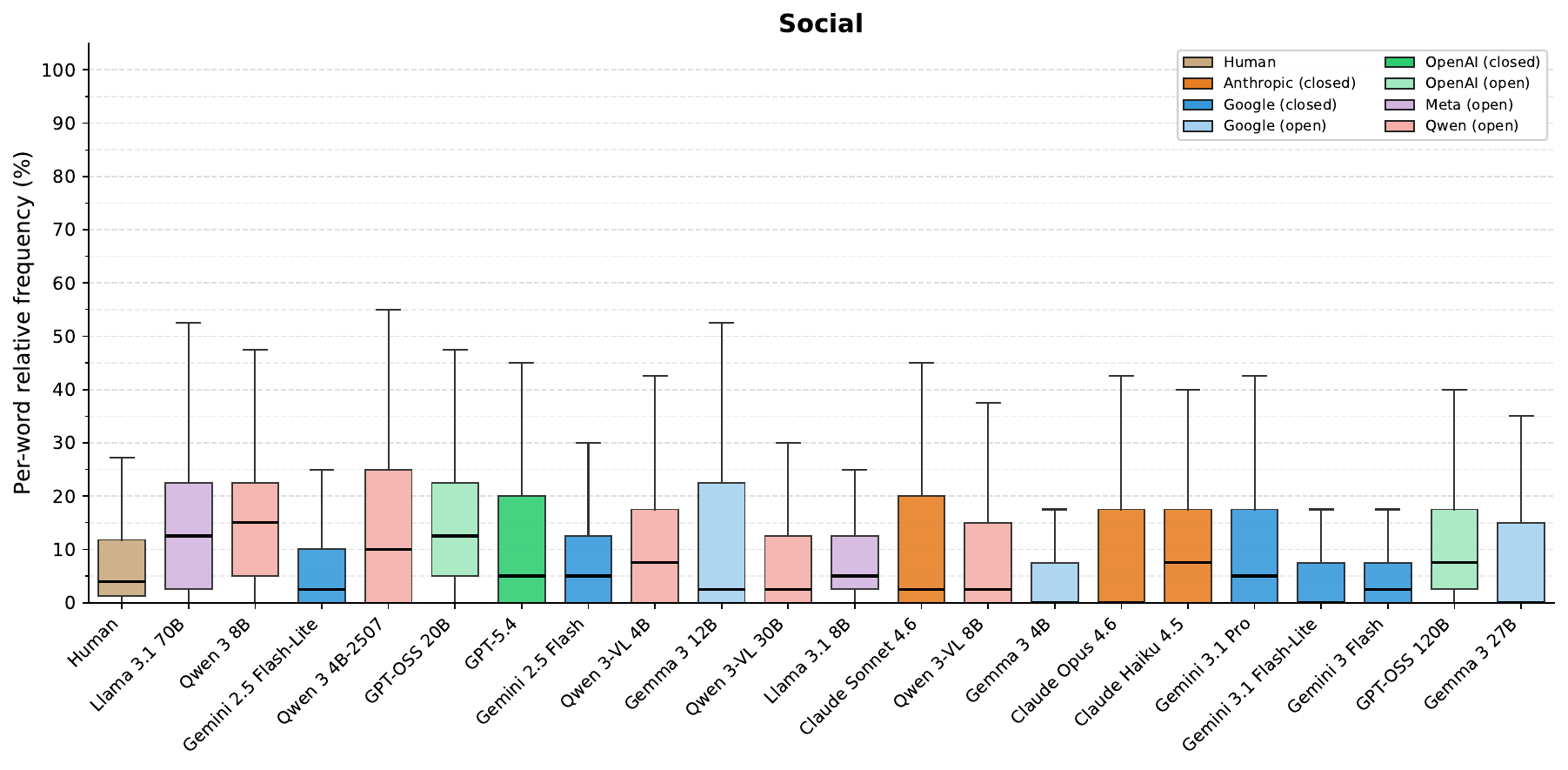}
        \vfill
        
        \includegraphics[width=0.8\textwidth]{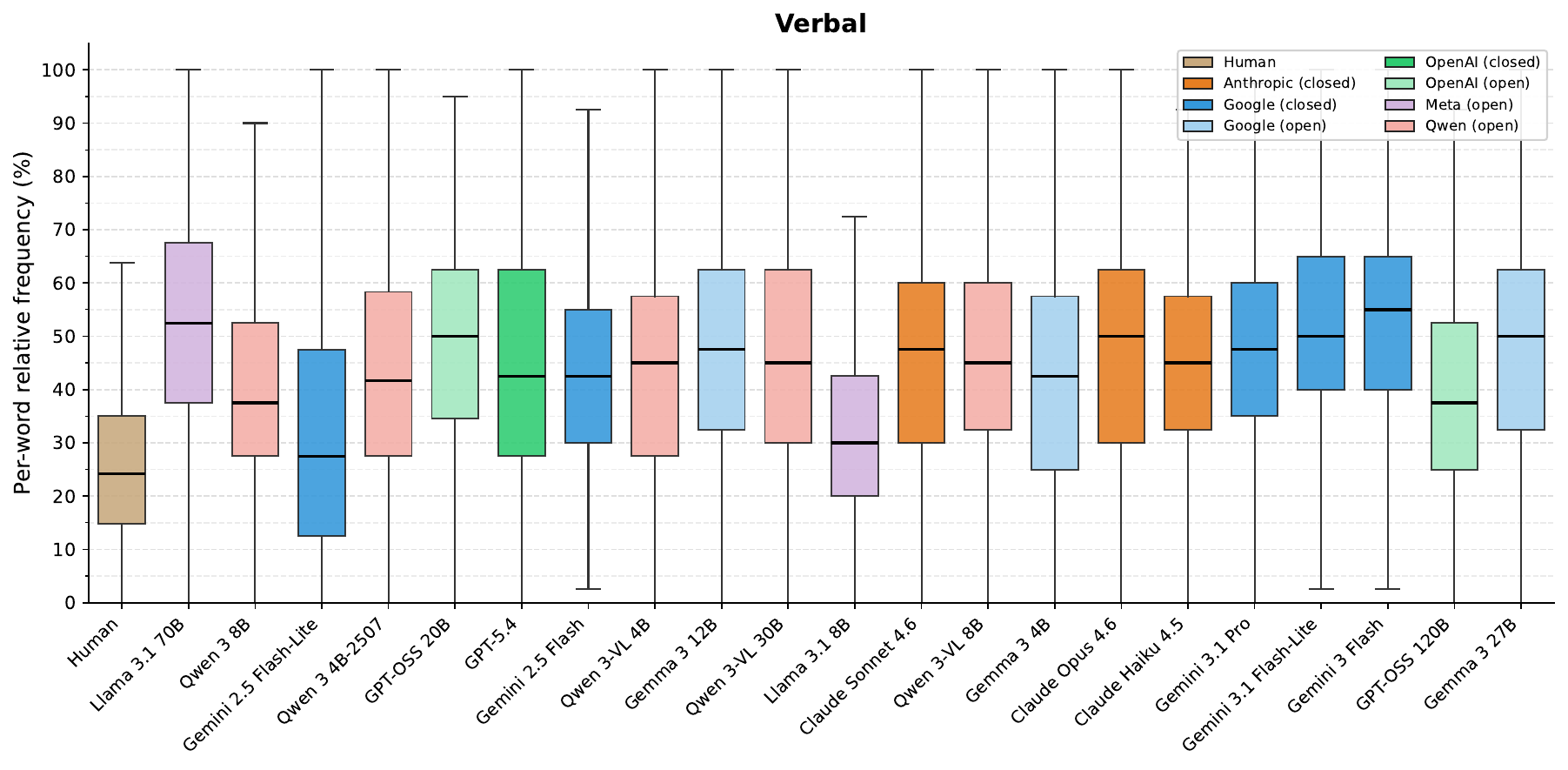}
        
        \caption{Property-frequency boxplots for all 21 LLMs, sorted by Mean~r.}
        \label{fig:boxplots-exp1}
    \end{minipage}
\end{figure}

Figure~\ref{fig:boxplots-exp1} presents the frequency profiles for all models. 
Table~\ref{tab:app-harpaintner-full} details the performance of all 21 models evaluated in our panel. Smaller models occasionally outperform substantially larger ones, suggesting that simply increasing parameter count does not organically close the grounding gap.

\begin{table}[ht]
  \caption{Full experiment~1 leaderboard, all 21 models in the panel, sorted by Mean~$r$. \fix{Ten generation runs per model.} Bold marks the column max.}
  \label{tab:app-harpaintner-full}
  \centering\footnotesize
  \smallskip
  \begin{tabular}{rlccccc}
    \toprule
    Rank & Model & Mean $r$ & Sensorimotor & Internal & Social & Verbal\\
    \midrule
     & \emph{Estimated human ceiling} & \emph{0.974} & \emph{0.966} & \emph{0.964} & \emph{0.971} & \emph{0.992} \\
    \midrule
     1 & Llama 3.1 70B            & \textbf{0.375} \fix{$\pm$ 0.011} & 0.344 & \textbf{0.464} & 0.424          & 0.267          \\
     2 & Qwen3 8B                 & 0.373 \fix{$\pm$ 0.015}          & \textbf{0.351}          & 0.390          & \textbf{0.462} & \textbf{0.288} \\
     3 & Gemini 2.5 Flash-Lite    & 0.348 \fix{$\pm$ 0.009}          & 0.276          & 0.382          & 0.456          & 0.278          \\
     4 & Qwen3 4B       & 0.314 \fix{$\pm$ 0.013}          & 0.335          & 0.405          & 0.318          & 0.200          \\
     5 & GPT-OSS 20B              & 0.302 \fix{$\pm$ 0.013}          & 0.326          & 0.255          & 0.429          & 0.196          \\
     6 & GPT-5.4                  & 0.301 \fix{$\pm$ 0.005}          & 0.332          & 0.258          & 0.400          & 0.213          \\
     7 & Gemini 2.5 Flash         & 0.296 \fix{$\pm$ 0.012}          & 0.291          & 0.310          & 0.364          & 0.218          \\
     8 & Qwen3-VL 4B              & 0.295 \fix{$\pm$ 0.007}          & 0.344          & 0.251          & 0.326          & 0.260          \\
     9 & Gemma-3-12B          & 0.292 \fix{$\pm$ 0.005}          & 0.277          & 0.283          & 0.357          & 0.251          \\
    10 & Qwen3-VL 30B             & 0.292 \fix{$\pm$ 0.007}          & 0.278          & 0.240          & 0.412          & 0.237          \\
    11 & Llama 3.1 8B             & 0.283 \fix{$\pm$ 0.014}          & 0.186          & 0.356          & 0.406          & 0.185          \\
    12 & Claude Sonnet 4.6        & 0.277 \fix{$\pm$ 0.006}          & 0.287          & 0.242          & 0.352          & 0.225          \\
    13 & Qwen3-VL 8B              & 0.273 \fix{$\pm$ 0.009}          & 0.313          & 0.256          & 0.325          & 0.199          \\
    14 & Gemma-3-4B          & 0.270 \fix{$\pm$ 0.006}          & 0.279          & 0.266          & 0.291          & 0.244          \\
    15 & Claude Opus 4.6          & 0.264 \fix{$\pm$ 0.004}          & 0.203          & 0.205          & 0.362          & 0.284          \\
    16 & Claude Haiku 4.5         & \fix{0.258} \fix{$\pm$ 0.008}          & \fix{0.250}          & \fix{0.196}          & 0.310          & \fix{0.277}          \\
    17 & Gemini 3.1 Pro   & 0.256 \fix{$\pm$ 0.008}          & 0.307          & 0.163          & 0.383          & 0.171          \\
    18 & Gemini 3.1 Flash-Lite    & 0.223 \fix{$\pm$ 0.007}          & 0.264          & 0.134          & 0.279          & 0.216          \\
    19 & Gemini 3 Flash   & 0.216 \fix{$\pm$ 0.013}          & 0.251          & 0.182          & 0.294          & 0.137          \\
    20 & GPT-OSS 120B             & 0.199 \fix{$\pm$ 0.010}          & 0.210          & 0.238          & 0.230          & 0.116          \\
    21 & Gemma-3-27B          & 0.196 \fix{$\pm$ 0.005}          & 0.262          & 0.115          & 0.262          & 0.144          \\
    \bottomrule
  \end{tabular}
  \par\smallskip
\end{table}

\section{Extended results for Experiment 2}
\label{app:exp2}

Here we present extended quantitative results for the second experiment (Section~\ref{sec:exp2}) based on the \citet{kelly2024conceptual} dataset. The Kelly stimulus set partitions the \fix{$354$} target words into three matched conditions: $118$ abstract emotion words, \fix{$117$} abstract non-emotion words, and $119$ concrete words. The main text reports alignment on the combined \fix{$235$}-word abstract subset (emotion plus non-emotion). For completeness, in this appendix we also report the leaderboard restricted to the $119$-word concrete subset, which serves as a control condition: concrete nouns are the easy case for property generation, since they admit clearly perceptual properties that all candidate models recover well.

Properties are classified by Gemini-2.5-Flash into the four superordinate Kelly categories: \textbf{Taxonomic}, \textbf{Entity}, \textbf{Situation}, and \textbf{Introspective}. 
Each model produces $10$ runs over the full \fix{$354$}-word stimulus set; per-concept category-frequency vectors are averaged across runs before correlating with the human reference distribution. The primary metric is \fix{Mean~$r$} between the per-concept distributions of the model and the human aggregate, computed separately on the abstract and concrete subsets.

\begin{figure}[t]
  \centering
  \includegraphics[width=0.82\linewidth]{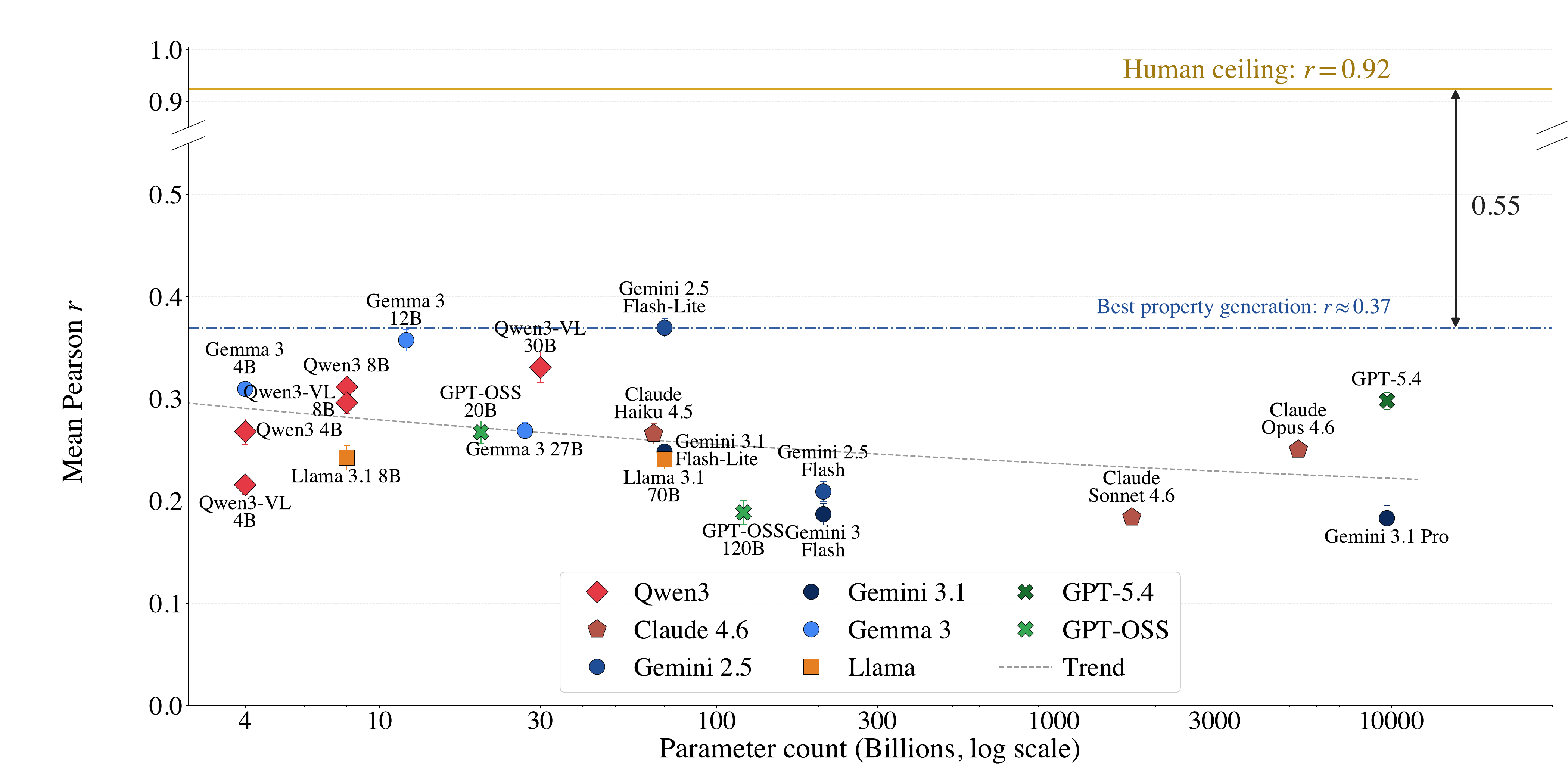}
  \caption{Mean r for all LLMs vs.\ the human ceiling on Experiment~2 \citep{kelly2024conceptual}. Parameter count for closed models is based on estimations from \citet{li2026incompressible}. \fix{The dashed curve shows the saturating fit to the 21 models (trend). The dash-dotted line marks the best model, and the arrow its distance to the human ceiling.}}
  \label{fig:exp2-scaling}
\end{figure}

\begin{figure}[t]
  \centering
  \includegraphics[width=0.5\linewidth]{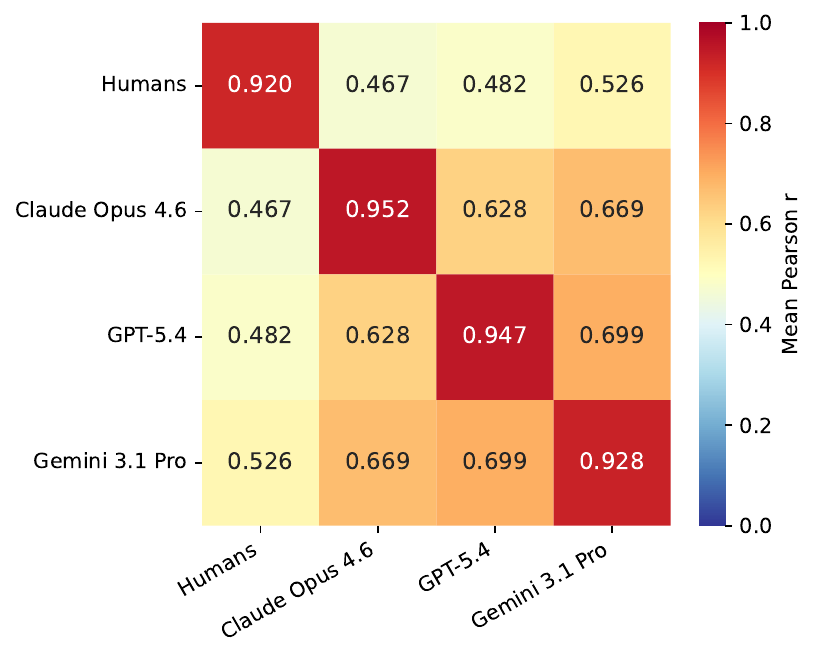}
  \caption{Experiment~2 properties correlation heatmap for the concrete word subset. The correlation between LLMs--humans is closer to LLMs-LLMs in this case, compared to Fig.~\ref{fig:inter_model_heatmaps}.}
  \label{fig:exp2-concrete-heatmap}
\end{figure}

Figure~\ref{fig:exp2-scaling} replicates the scaling plot for the Experiment 2 setup: even the strongest models reach only \fix{Mean~$r \approx 0.37$} on the abstract subset, far below the human-to-human ceiling of \fix{$r \approx 0.92$}, and there is no monotonic trend in parameter count. Table~\ref{tab:exp2-leaderboard-full} lists every model on both subsets sorted by abstract \fix{Mean~$r$}. The pattern mirrors Experiment~1: the abstract-subset gap remains roughly flat across model sizes and providers, while the same models reach much higher alignment ($r \in$ \fix{$[0.36, 0.53]$}) on the concrete subset. This contrast is consistent with the central claim of the paper: the gap is specific to abstract concepts, not a general property-generation deficit. Figure~\ref{fig:exp2-concrete-heatmap} confirms this from a different angle: on the concrete subset the human-to-model and model-to-model correlations sit in similar ranges, whereas on the abstract subset (Fig.~\ref{fig:inter_model_heatmaps} in the main text) models are much more correlated with each other than with humans.

\begin{table}[t]
  \caption{Extended leaderboard on experiment~2 \citet{kelly2024conceptual}, for both the abstract and concrete subsets. Models are sorted by \fix{Mean $r$} on the abstract subset.}
  \label{tab:exp2-leaderboard-full}
  \centering\small
  \begin{tabular}{lcc}
    \toprule
    Model & Abstract & Concrete \\
    \midrule
    \emph{Human ceiling} & \fix{\emph{0.924 $\pm$ 0.006}} & \emph{0.920 $\pm$ 0.008} \\
    \midrule
    Gemini 2.5 Flash-Lite & \textbf{\fix{0.370 $\pm$ 0.009}} & 0.488 $\pm$ 0.014 \\
    Gemma 3 12B & \fix{0.357 $\pm$ 0.011} & 0.393 $\pm$ 0.010 \\
    Qwen 3-VL 30B & \fix{0.331 $\pm$ 0.014} & \fix{0.480 $\pm$ 0.015} \\
    Qwen 3 8B & \fix{0.312 $\pm$ 0.008} & 0.417 $\pm$ 0.016 \\
    Gemma 3 4B & \fix{0.310 $\pm$ 0.008} & 0.423 $\pm$ 0.015 \\
    GPT-5.4 & \fix{0.298 $\pm$ 0.009} & 0.482 $\pm$ 0.012 \\
    Qwen 3-VL 8B & \fix{0.296 $\pm$ 0.010} & \fix{0.393 $\pm$ 0.015} \\
    Gemma 3 27B & \fix{0.269 $\pm$ 0.008} & 0.447 $\pm$ 0.016 \\
    Qwen 3 4B & \fix{0.268 $\pm$ 0.013} & \fix{0.360 $\pm$ 0.025} \\
    GPT-OSS 20B & \fix{0.267 $\pm$ 0.012} & \fix{0.422 $\pm$ 0.028} \\
    Claude Haiku 4.5 & \fix{0.266 $\pm$ 0.010} & \fix{0.389 $\pm$ 0.017} \\
    Claude Opus 4.6 & \fix{0.251 $\pm$ 0.007} & 0.467 $\pm$ 0.013 \\
    Gemini 3.1 Flash-Lite & \fix{0.248 $\pm$ 0.007} & 0.465 $\pm$ 0.012 \\
    Llama 3.1 8B & \fix{0.243 $\pm$ 0.012} & \fix{0.449 $\pm$ 0.022} \\
    Llama 3.1 70B & \fix{0.241 $\pm$ 0.009} & 0.469 $\pm$ 0.017 \\
    Qwen 3-VL 4B & \fix{0.216 $\pm$ 0.010} & \fix{0.391 $\pm$ 0.017} \\
    Gemini 2.5 Flash & \fix{0.209 $\pm$ 0.009} & \textbf{0.531 $\pm$ 0.018} \\
    GPT-OSS 120B & \fix{0.189 $\pm$ 0.012} & 0.401 $\pm$ 0.021 \\
    Gemini 3 Flash & \fix{0.187 $\pm$ 0.010} & \fix{0.463 $\pm$ 0.022} \\
    Claude Sonnet 4.6 & \fix{0.184 $\pm$ 0.007} & 0.436 $\pm$ 0.013 \\
    Gemini 3.1 Pro & \fix{0.183 $\pm$ 0.012} & \fix{0.526 $\pm$ 0.018} \\
    \bottomrule
  \end{tabular}
  \par\smallskip
\end{table}

\section{Rating experiment additional results}
\label{app:exp-ratings}

\paragraph{Task.}
We replicate the rating study of \citet{troche2017} using $751$ English nouns rated on $14$ \fix{cognitive} dimensions under the uniform framing \emph{``I relate this word to [X]''}, with responses on a $1$ to $7$ Likert scale ($1$ = Strongly Disagree, $7$ = Strongly Agree). Each model produces ten rating seeds; per-word ratings are averaged across seeds before correlating with the human norms. The primary metric is the Pearson~$r$ averaged over the $14$ dimensions. \fix{The} $14$ dimensions are organised into \fix{five} components: \textbf{Sensory} (Color, Taste/Smell, Tactile, Visual Form, Auditory), \fix{\textbf{Motor} (Self-Motion),} \textbf{Internal} (Emotion, Polarity, Morality, Thought), \fix{\textbf{Social} (Social),} and \textbf{Magnitude} (Space, Quantity, Time). Table~\ref{tab:rating-exp-leaderboard-full} reports the per-component Pearson correlations alongside the overall Mean~$r$ for every model out of the $21$.

\paragraph{Results.}
Three patterns stand out in Table~\ref{tab:rating-exp-leaderboard-full}, and each one provides a different angle on the grounding gap reported in the main text.

The first observation is that the strongest models are \fix{close to} the human ceiling on this task. Claude Sonnet~4.6 reaches Mean~$r = 0.760$ against a cross-dimension human ceiling of approximately \fix{$0.87$} (Appendix~\ref{app:ceilings}). The same models top out at \fix{$r \approx 0.38$} on the property generation task of Experiment~1 against a ceiling of $0.97$ (Table~\ref{tab:app-harpaintner-full}), \fix{a gap about five times larger} to the human reference. We take the contrast as direct evidence for the central claim of the paper: when a grounding dimension is named explicitly in the prompt, current LLMs can recover it; the gap is therefore specific to the spontaneous property generation setting and is not a general inability to register sensorimotor, internal, or social content.

The second observation is that scaling helps recognition, but only on a subset of the components. Within \fix{most} open-weight families we tested, the Sensory component improves dramatically with model size: Gemma scales from $-0.086$ at $4$B to $0.704$ at $27$B, Llama scales from $-0.161$ at $8$B to $0.729$ at $70$B, and Qwen3 scales from $0.321$ at $4$B to $0.807$ at $8$B (Table~\ref{tab:rating-exp-leaderboard-full}). The Magnitude component \fix{also increases} within each family. The Internal and Social components, by contrast, are essentially flat across model sizes: every model in Table~\ref{tab:rating-exp-leaderboard-full} sits in the range $[0.58, 0.84]$ on these two components, regardless of parameter count or provider. Recognition of affective and social content is therefore not a capability bottleneck at any scale we tested. Perceptual grounding, on the other hand, is recoverable only at sufficient scale, and once that scale is reached the model can rate perceptual content as well as humans do.

The third observation is that the Motor sub-dimension is a persistent weakness that scaling does not close. As shown in Table~\ref{tab:rating-exp-leaderboard-full}, \fix{Motor scores are weaker overall, and some are very weak}, with Gemini~3.1~Pro at $0.119$, Gemini~3~Flash at $0.189$, and Gemini~3.1~Flash-Lite even falling below zero at $-0.058$. More importantly, the within-family scaling that lifts Sensory does not lift Motor: Gemma improves only modestly from $0.491$ at $4$B to $0.614$ at $27$B (with $12$B sitting higher than $27$B at $0.693$), Llama actually drops from $0.528$ at $8$B to $0.495$ at $70$B, and Qwen3 drops from $0.540$ at $4$B to $0.375$ at $8$B. 

\begin{table}[t]
\centering
\small
\caption{Rating~experiment results for all 21 models. Mean~$r$ is the average of the 14 per-dimension Pearson correlations against the human ratings.}
\label{tab:rating-exp-leaderboard-full}
\setlength{\tabcolsep}{5pt}
\begin{tabular}{lcccccc}
\toprule
Model & Mean $r$ & Sensory & Motor & Internal & Social & Magnitude \\
\midrule
Claude Sonnet 4.6 & 0.760 & 0.839 & 0.583 & 0.787 & 0.827 & 0.628 \\
Claude Opus 4.6 & 0.755 & 0.868 & 0.393 & 0.779 & 0.812 & 0.637 \\
Gemini 3 Flash & 0.730 & 0.889 & 0.189 & 0.750 & 0.793 & 0.600 \\
Gemini 2.5 Flash & 0.720 & 0.872 & 0.229 & 0.740 & 0.813 & 0.574 \\
Gemini 3.1 Pro & 0.719 & 0.883 & 0.119 & 0.753 & 0.790 & 0.576 \\
GPT-5.4 & 0.717 & 0.846 & 0.432 & 0.733 & 0.737 & 0.567 \\
Llama 3.1 70B & 0.707 & 0.729 & 0.495 & 0.754 & 0.832 & 0.637 \\
Gemma 3 27B & 0.706 & 0.704 & 0.614 & 0.775 & 0.829 & 0.606 \\
Qwen 3 8B & 0.703 & 0.807 & 0.375 & 0.717 & 0.828 & 0.580 \\
GPT-OSS 20B & 0.701 & 0.757 & 0.513 & 0.712 & 0.825 & 0.613 \\
GPT-OSS 120B & 0.695 & 0.709 & 0.512 & 0.745 & 0.837 & 0.617 \\
Gemini 2.5 Flash-Lite & 0.679 & 0.719 & 0.434 & 0.743 & 0.787 & 0.572 \\
Gemini 3.1 Flash-Lite & 0.677 & 0.829 & -0.058 & 0.686 & 0.794 & 0.618 \\
Claude Haiku 4.5 & 0.672 & 0.801 & 0.071 & 0.711 & 0.768 & 0.571 \\
Qwen 3-VL 30B & 0.665 & 0.641 & 0.604 & 0.712 & 0.827 & 0.607 \\
Gemma 3 12B & 0.559 & 0.326 & 0.693 & 0.758 & 0.811 & 0.554 \\
Qwen 3-VL 8B & 0.517 & 0.300 & 0.557 & 0.693 & 0.764 & 0.551 \\
Qwen 3 4B-2507 & 0.507 & 0.321 & 0.540 & 0.687 & 0.703 & 0.499 \\
Qwen 3-VL 4B & 0.457 & 0.192 & 0.539 & 0.682 & 0.740 & 0.479 \\
Llama 3.1 8B & 0.341 & -0.161 & 0.528 & 0.736 & 0.710 & 0.464 \\
Gemma 3 4B & 0.297 & -0.086 & 0.491 & 0.588 & 0.580 & 0.391 \\
\bottomrule
\end{tabular}
\smallskip

\end{table}

\section{Variance and uncertainty in Experiment~1}
\label{sec:appendix-variance}

The Experiment~1 leaderboard reports each model's Mean
Pearson $r$ as the point estimate from $10$ independent generation runs,
plus a bootstrap standard deviation. Two questions a reader is right to
ask: (i) is a $10$-run point estimate close to the asymptote that the
same model would reach with infinitely many runs, and (ii) what does the
bootstrap $\pm$ bar actually quantify? We answer both empirically by
running \textsc{Gemma-3-4B-IT} 100 times, treating that aggregate as the ground truth, and comparing it to
what we would have estimated from random $10$-run subsets of the same
data. 

\subsection{Convergence: how accurate is the 10-run estimate?}
\label{sec:appendix-variance-convergence}

\begin{figure}[t]
\centering
\includegraphics[width=0.7\linewidth]{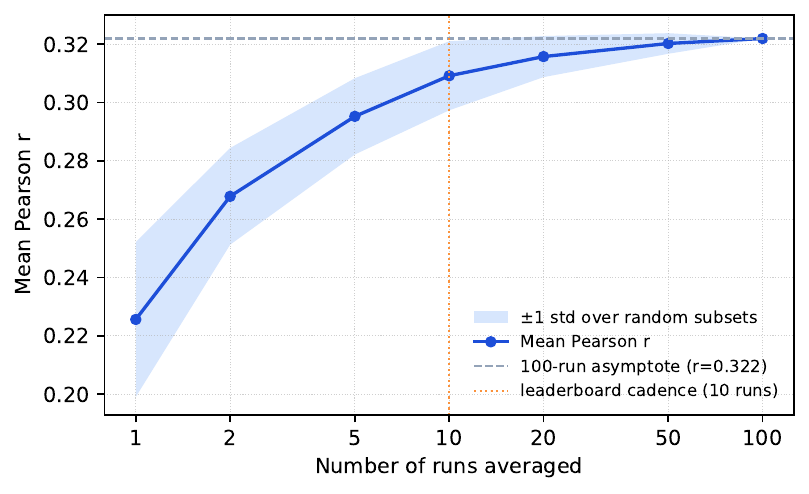}
\caption{Mean Pearson $r$ vs.\ number of runs averaged, on
Experiment 1, for \textsc{Gemma-3-4B-IT} coded by
Gemini-2.5-Flash-Lite. Solid line: mean over $50$ random subsets at each
subset size; band: $\pm 1$~std across subsets; dashed: 100-run
asymptote; dotted: the $10$-run cadence used by the Experiment~1 leaderboard.}
\label{fig:appendix-variance-convergence}
\end{figure}

Figure~\ref{fig:appendix-variance-convergence} plots Mean $r$ as a
function of the number of runs averaged, with shaded $\pm 1$~std bands
over $50$ random subsets at each subset size. The estimate climbs
monotonically toward the 100-run asymptote of
$r = +0.322$. At the Experiment~1 leaderboard cadence of $10$
runs, the average estimate is $r = +0.309$, lower than
the asymptote by $\Delta r = +0.013$. 

The curve has two practical implications. \emph{For absolute $r$ values
on a single model}, a $10$-run point estimate sits about $0.01$ below
the asymptote it would converge to, so quoted $r$ values are slightly
conservative. \emph{For ranking models against each other}, this bias
is approximately constant across models, so the relative ordering on the $10$-run
Experiment~1 leaderboard is stable.

\subsection{Bootstrap calibration}
\label{sec:appendix-variance-bootstrap}

The Experiment~1 leaderboard's $\pm$ value next to each model's Mean $r$ is a
bootstrap standard deviation computed from $N_{\text{boot}} =
1000$ resamples of the $10$ runs with replacement. For each
resample we re-aggregate the per-word frequencies, recompute Mean $r$
against the human norms, and take the standard deviation of the
resulting distribution. 

\section{Estimation of human correlation ceilings}
\label{app:ceilings}

Each experiment reports a per-category Pearson~$r$ human ceiling, defined as the correlation between two independent panels of $N$ raters scored on the same stimuli. This is the largest $r$ a noiseless model could expect to reach against the published norms, which are themselves panel means. We obtain it in one of two ways depending on what each original study released. \citet{kelly2024conceptual} provides participant-level responses, so we compute the ceiling by direct resampling. \citet{harpaintner2018semantic} and \citet{troche2017} release only per-word means and SDs, so we estimate the ceiling from those aggregates using the standard ICC$+$Spearman--Brown chain: $\mathrm{ICC}_1$ recovers single-rater reliability from the ratio of between-word to total variance, and Spearman--Brown projects single-rater reliability up to the $N$-vs-$N$ scale via $r_{\text{full}} = 2 r_{\text{split}}/(1 + r_{\text{split}})$. We always report $r_{\text{full}}$, because the human norm against which models are correlated is a full-panel mean rather than a single rater.

\subsection{Experiment~2: empirical split-half}
\label{app:ceilings-kelly}

We draw $1000$ random half-splits of the participants per concept (median $N{\approx}30$ raters across \fix{$354$} concepts), compute per-concept property-category percentages on each half, and average Pearson~$r$ across splits. Spearman--Brown then projects the half-panel correlation to \fix{$r_{\text{full}} = 0.924$ on the 235-word} abstract subset used in the main text and \fix{$r_{\text{full}} = 0.920$} on the 119-word concrete subset used in the appendix.


\subsection{Validation of the ceiling estimation}
\label{app:ceilings-validation}

Experiment~2 is the only dataset where both methods are available on the same data, since the participant-level release also yields per-word means and SDs. Applying the ICC$+$SB chain we use for the other experiments to the Experiment~2 category-frequency matrix gives a within-dataset cross-check on the estimate.

\begin{table}[ht]
  \caption{ICC$+$Spearman--Brown analytical estimate vs.\ $1000$-split empirical ceiling on Experiment~2, both at $r_{\text{full}}$ scale.}
  \label{tab:app-kelly-iccsb-check}
  \centering\small
  \begin{tabular}{lccc}
    \toprule
    Slice & ICC$+$SB & Empirical & $|\Delta|$ \\
    \midrule
    \fix{Abstract (235 words)} & \fix{0.941} & \fix{0.924} & \fix{0.017} \\
    Concrete (119 words) & \fix{0.929} & \fix{0.920} & \fix{0.009} \\
    \bottomrule
  \end{tabular}
\end{table}

The two methods agree to within \fix{$0.02$} on both slices, with the analytical estimate sitting fractionally above the empirical ceiling. We therefore consider the ICC$+$SB ceilings reported below as accurate to within \fix{${\sim}0.02$}.

\paragraph{Experiment~1 ceilings.}
\citet{harpaintner2018semantic} publish per-word frequencies in four aggregate categories. The resulting $r_{\text{full}}$ ceilings are:
\begin{equation*}
  \text{Sensorimotor: } 0.966, \quad
  \text{Internal: } 0.964, \quad
  \text{Social: } 0.971, \quad
  \text{Verbal: } 0.992 \quad (\overline{r}_{\text{full}} = 0.974).
\end{equation*}


\paragraph{Rating-experiment ceilings.}
\citet{troche2017} publish per-word means and SDs on each of the 14 grounding dimensions. We apply ICC$+$SB dimension-by-dimension with $\sigma^2_{\text{between}}$ from the cross-stimulus variance of per-word means, $\sigma^2_{\text{within}}$ from the mean squared SD, and $N$ from the median raters per word ($\approx 24$). The 14 $r_{\text{full}}$ ceilings range from \fix{$0.75$ on Space} to \fix{$0.96$} on Visual Form (cross-dimension mean \fix{$0.87$}).


\section{Prompting setup}
\label{app:prompts}
This section presents verbatim \fix{the prompts} used on the three experiments. The literal token \texttt{\{word\}} is substituted by each target concept on each call, and \texttt{\{properties\}} (in the coding prompts) by the comma-separated list of generated properties for that concept.

\subsection{Property Generation Prompt (Experiment 1)}
\label{app:prompts-gen-harpaintner}

The generation prompt for Experiment~1 is shown below in Figure~\ref{fig:appendix-prompt-harpaintner}. It asks the model for exactly four spontaneous properties per noun, with one in-context example (\textit{sympathy}). It follows the original property listing instructions of \citet{harpaintner2018semantic}. 

\begin{figure}[ht]
\centering
\includegraphics[width=\linewidth]{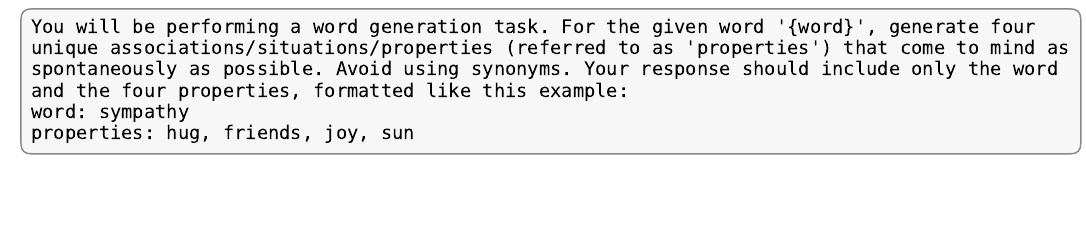}
\caption{Property generation prompt for Experiment~1 \citet{harpaintner2018semantic}, reproduced verbatim. The placeholder \texttt{\{word\}} is replaced by each of the target nouns at inference time.}
\label{fig:appendix-prompt-harpaintner}
\end{figure}

\subsection{Property Generation Prompt (Experiment 2)}
\label{app:prompts-gen-kelly}

The generation prompt for Experiment~2 is shown below in Figure~\ref{fig:appendix-prompt-kelly-gen}. It asks the model for exactly five spontaneous clues per noun, with two in-context examples (\textit{dog} and \textit{dejected}). It follows the human rater instructions in Appendix~A of \citet{kelly2024conceptual}. 



\begin{figure}[ht]
\centering
\includegraphics[width=\linewidth]{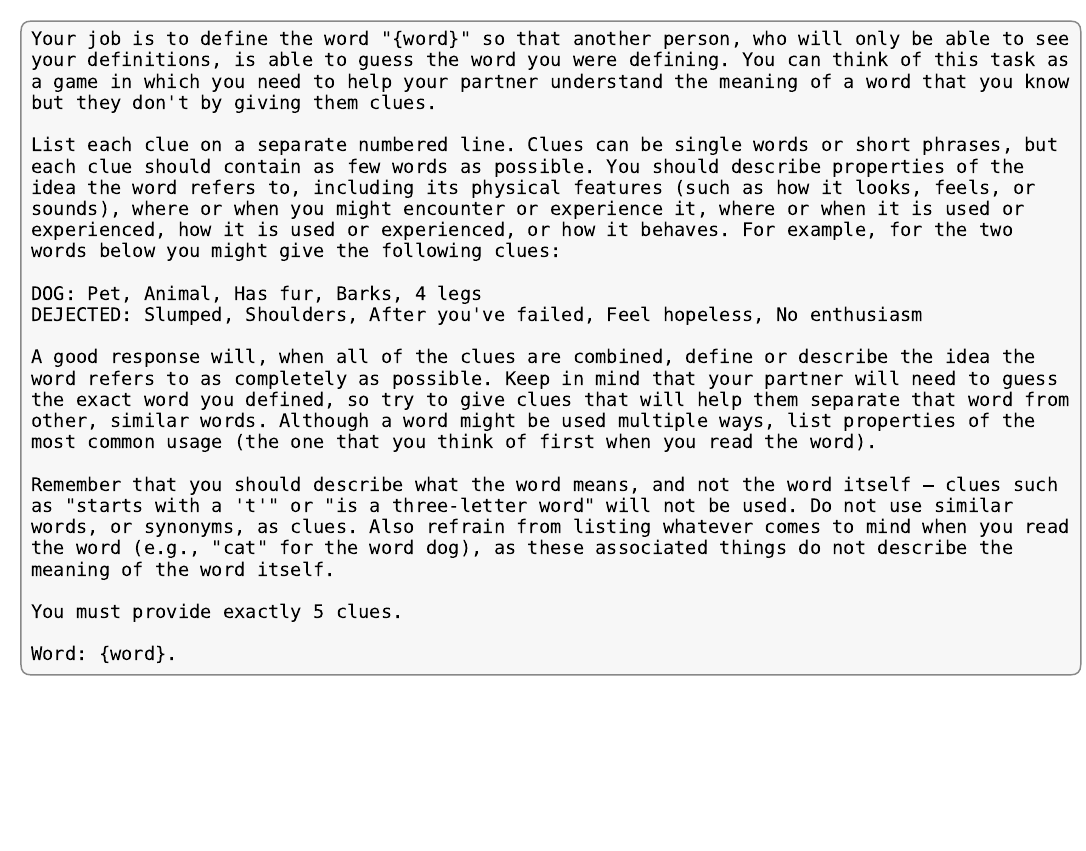}
\caption{Property generation prompt used for Experiment~2 \citet{kelly2024conceptual}, reproduced verbatim. The placeholder \texttt{\{word\}} is replaced by each target concept at inference time.}
\label{fig:appendix-prompt-kelly-gen}
\end{figure}

\subsection{Coding Prompt (Experiment 1)}
\label{app:prompts-cod-harpaintner}

\begin{figure}[ht]
\centering
\includegraphics[width=\linewidth]{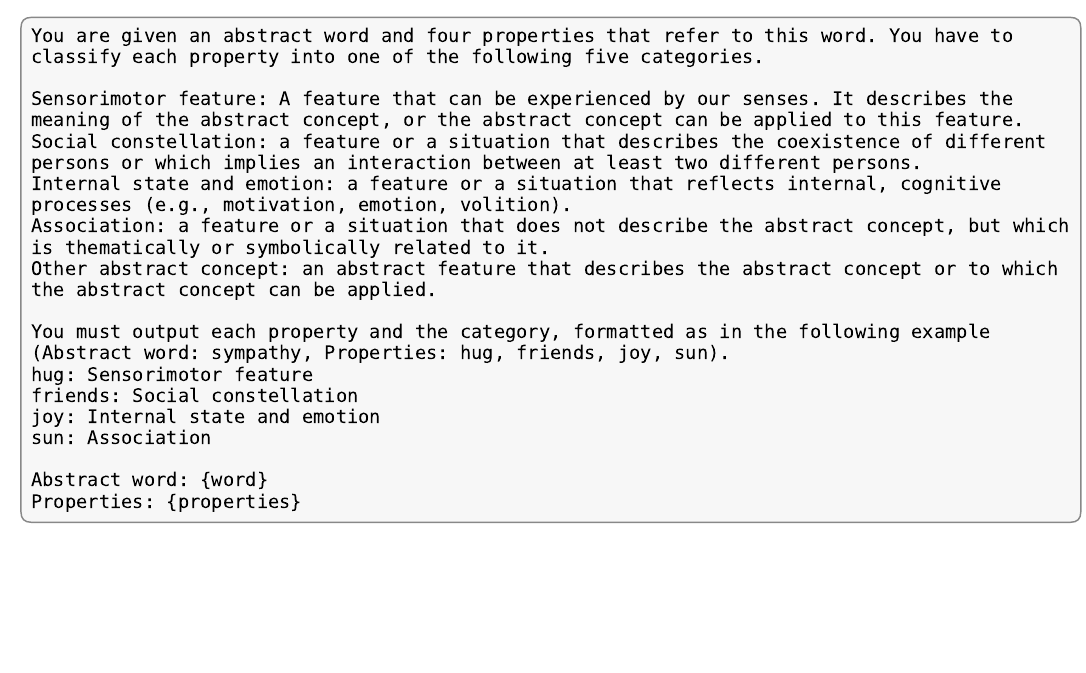}
\caption{Coding prompt for Experiment~1. Categories are listed verbatim from \citet{harpaintner2018semantic}; \texttt{\{word\}} and \texttt{\{properties\}} are replaced at inference time by the concept and the comma-separated generated property list.}
\label{fig:appendix-prompt-harpaintner-coding}
\end{figure}

Each generated property is assigned one of five mutually exclusive categories: \textbf{Sensorimotor feature} (a feature experienced by the senses); \textbf{Social constellation} (the coexistence or interaction of persons); \textbf{Internal state and emotion} (internal cognitive processes such as motivation, emotion, volition); \textbf{Association} (thematically or symbolically related but not descriptive); and \textbf{Other abstract concept} (an abstract feature that describes the concept). After coding, the two abstract categories: \textbf{Association} and \textbf{Other abstract concept} are merged to form the final abstract category \textbf{Verbal association}. The prompt provides a single in-context example (the \textit{sympathy} properties from the generation prompt, one per category) and asks the coder to emit one labelled line per property. The full coding prompt is shown in Figure~\ref{fig:appendix-prompt-harpaintner-coding}.

\subsection{Rating Experiment Prompt}
\label{app:prompts-troche}

\begin{figure}[ht]
\centering
\includegraphics[width=\linewidth]{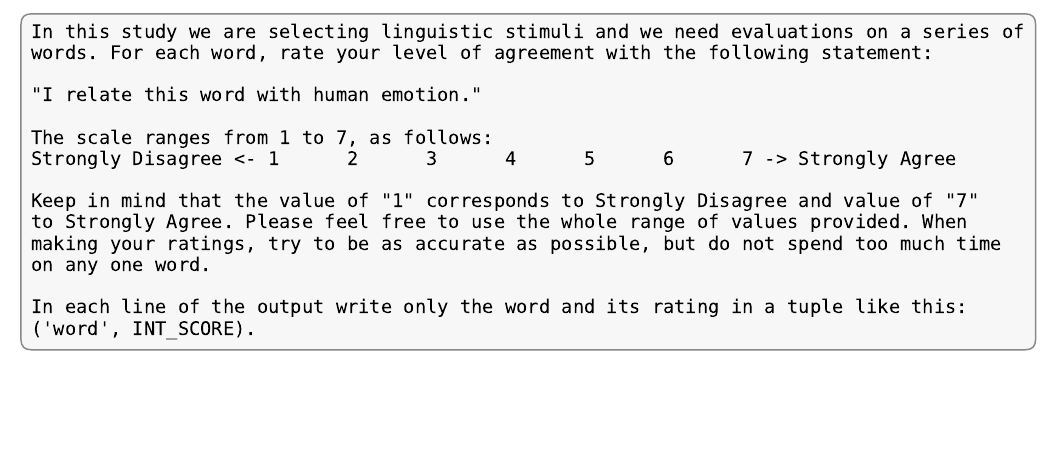}
\caption{Rating prompt for the Troche benchmark, shown for the \textit{emotion} dimension. The same instruction skeleton is reused for all 14 dimensions; only the target sentence (``I relate this word with \textit{X}.'') changes between dimensions, with the wording for each dimension taken verbatim from \citet{troche2017}.}
\label{fig:appendix-prompt-troche}
\end{figure}

The rating experiment \citet{troche2017} uses 14 \fix{cognitive} dimensions: \textit{color, taste/smell, tactile, visual form, auditory, emotion, polarity, social, morality, thought, self-motion, space, quantity, time}. Every dimension uses the same instruction skeleton, with only the dimension specific statement substituted in. The full prompt for the \textit{emotion} dimension is shown verbatim in Figure~\ref{fig:appendix-prompt-troche}; the other thirteen dimensions are produced by replacing the sentence ``I relate this word with human emotion.'' with the analogous sentence for the target dimension, taken verbatim from \citet{troche2017}. The output is a list of \texttt{('word', INT\_SCORE)} tuples on a 1--7 Likert scale.

\clearpage

\section{Extended mechanistic analysis results}
\label{app:mech-analysis}

\subsection{Feature identification dataset construction}
\label{app:feat-identification}

This appendix documents how we built the \fix{$425$}-noun balanced corpus used as the candidate pool for SAE feature identification (Section~\ref{sec:feature-identify}). Published norm sets cover individual dimensions of abstract-concept structure but no single resource gives us four balanced experiential categories on the same vocabulary, as well as \citet{troche2017} does. But we did not want to use our test-set to discover features, so we had to look at other norms. Additionally, we wanted to use only the very highest words per category, and this meant that we needed a larger dataset to subsample only the highest ones. We therefore built our own noun corpus by combining four complementary datasets. Each category is anchored to one primary norm and a single thresholded metric as shown in Table~\ref{tab:appendix-sae-feature-identification-norms} below, so that only the highest-scoring words per category are selected. A word that is not a noun or that satisfies thresholds for more than one grounded category is dropped from both. The final corpus comprises 110 sensorimotor, \fix{101} internal state, 102 social and 112 abstract nouns, for a total of \fix{425} nouns.

\begin{table}[htbp]
  \centering\small
  \caption{Norm sources and the metric extracted from each.}
  \label{tab:appendix-sae-feature-identification-norms}
  \begin{tabular}{lll}
    \toprule
    Norm & Source & Metric used \\
    \midrule
    Lancaster Sensorimotor & \citet{lynott2020lancaster}       & Max across \fix{$11$} modalities \\
    Diveica Socialness     & \citet{diveica2023quantifying}    & Socialness $1$--$7$ \\
    Glasgow Norms          & \citet{scott2019glasgow}          & Arousal $+$ $|$Valence$-5|$ \\
    Brysbaert Concreteness & \citet{brysbaert2014concreteness} & Concreteness $1$--$5$ \\
    \bottomrule
  \end{tabular}

  \vspace{2em} 
  \caption{Feature-identification corpus composition. Sensorimotor is further partitioned into $11$ Lancaster sub-modalities at $10$ words each (Visual, Auditory, Olfactory, Gustatory, Haptic, Interoception, Mouth, Head, Hand\_arm, Foot\_leg, Torso), enabling per-submodality stratification at the splitting stage.}
  \label{tab:appendix-sae-feature-identification-counts}
  \begin{tabular}{lrl}
    \toprule
    Category & $N$ & Primary source \\
    \midrule
    Sensorimotor      &   $110$ & Lancaster Norms ($11$ submodalities $\times$ $10$) \\
    Social            &   $102$ & Diveica Socialness \\
    Internal State    &   \fix{$101$} & Glasgow \\
    Abstract Baseline &   $112$ & Brysbaert (low concreteness) \\
    \bottomrule
  \end{tabular}
  \smallskip

\end{table}

\paragraph{Sentence generation}
\label{sec:appendix-sae-feature-identification-sentences}
Each of the \fix{$425$} nouns is paired with 10 context sentences that elicit the noun as the next token. Sentences are produced with Gemini-$3.1$-pro-preview at temperature $0$. For example, for $w{=}$\emph{surprise} one of the generated sentences is
\begin{quote}\itshape
``If you are feeling sad because you think everyone forgot your birthday, but then they jump out of the dark to make you smile, they have planned a \_\_\_''.
\end{quote}
The blank is the position where the target noun was originally written.

The prompt instructs Gemini to write $10$ numbered sentences ending with the target word, framed as a one-line riddle that makes the missing word easy to recover, and themed by the noun's category:
\begin{quote}\footnotesize\itshape
Write $10$ numbered sentences about everyday life that end with the word `\texttt{<WORD>}'. Your goal in writing these sentences is to make sure that it is obvious that the sentence ends with this word, like a very simple and easy riddle to learn the word. You must also make sure that the sentence is aligned with the theme of `\texttt{<THEME>}'. Start your answer with `$1$. \texttt{<the first sentence>}'.
\end{quote}

The string for \texttt{<THEME>} is filled per subcategory (for Sensorimotor: Visual~$\to$~``visual perception'', Auditory~$\to$~``auditory perception'', etc.) or per category (Internal, Social). For Abstract nouns the theme clause is dropped: there is no grounded modality to anchor on, so we let Gemini choose any framing. 

A regex post-processor then strips the target noun from the end of each generated line and verifies it is the same as the intended target.
The final \fix{$4{,}250$} successful rows generated for the \fix{$425$} training nouns are released publicly.


\paragraph{Last-word-target probing: per-noun activation signatures}
\label{sec:appendix-sae-feature-identification-probing}
Once the dataset is built, for each sentence in $S(w)$, we remove the target noun $w$ and feed the resulting prefix to the model. The SAE activation is read at the last token of the prefix, i.e.\ the position where the model is about to emit $w$ given everything before it. Per (layer, feature) we take the median across the $10$ sentences. Any feature that has a non-zero median for five or more nouns from a single category is considered as a candidate for that category.

\paragraph{Why this construction.}
The setup ensures (i)~The SAE sees a natural mid-sentence residual and therefore stays inside its training distribution, unlike isolated-token probes where many SAE features go silent. (ii)~The residual at position $n{-}1$ is a read-out of the pre-next-token distribution conditioned on the preceding context, which is target-word-specific without requiring the model to actually emit the target. (iii)~The aggregation is median over $10$ diverse context sentences, which suppresses any single sentence's effect.



\begin{figure}[ht]
\centering
\includegraphics[width=0.9\linewidth]{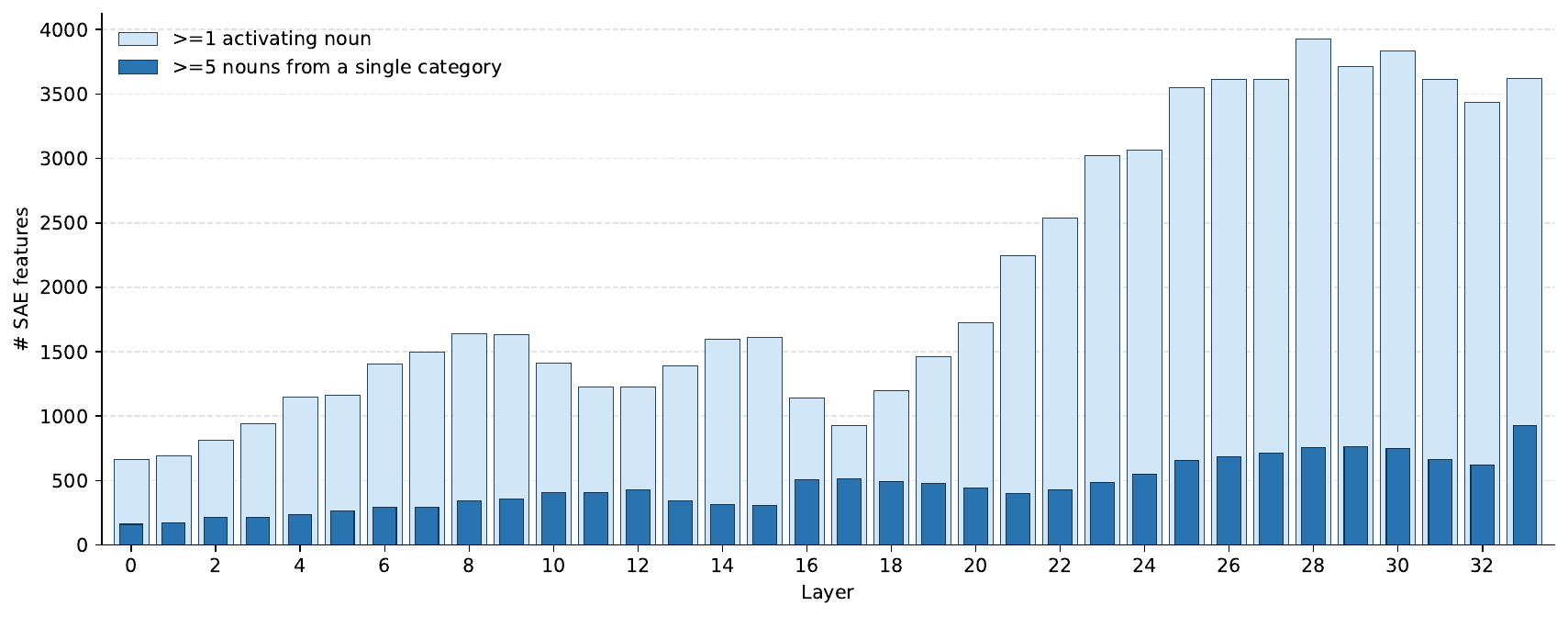}
\caption{Number of SAE features per layer identified with our detection algorithm in Gemma-3-4B.}
\label{fig:appendix-sae-feature-layers}
\end{figure}

\paragraph{Identified features per layer.}
In Figure~\ref{fig:appendix-sae-feature-layers} we see the features that have at least one non-zero median noun activation in our dataset and the ones that have at least five activations from a single category: the candidates. The features that have only a single activation are of course substantially more: either they fired randomly, or they are very specific to that noun. We observe an interesting phenomenon: until layers 15-17 the model seems to be encoding mostly and understanding the context of the previous and current tokens, so it does not have many features that correspond to the next token content which is our interest. From layer 17 (where it has about 1500 features) and after we see that the number of specific features increases rapidly and at layer 24 it reaches ~3500 features and stays constant, indicating that the model starts making decisions about the next word. On the other hand, the candidate features we selected exhibit a different behavior: they increase slightly but noticeably after layer 15 and then they remain constant, until a sudden step increase in the last layer. We hypothesize that this is because these features are mostly responsible for category-level decisions that have to do with the theme and context of the prefix, as intended by their selection process, and not so much for specific decisions dictated by linguistic or reasoning rules.

\subsection{Feature validation}
\label{app:feat-steer}


\paragraph{Feature interpretation.}
Determining the precise functionality of the features highly correlated with grounding dimensions remains challenging, as most lack an obvious semantic meaning. To facilitate interpretation and visualization, we provide links to Neuronpedia for representative features within each category. Additionally, we report automatic interpretations derived from our dataset activations using Claude-Opus-4.6. Note that the highest-correlated features discovered in our analysis are often not hosted on Neuronpedia, as the platform currently only supports a subset of layers (i.e., layers 9, 16, 22, and 29). Consequently, we present the best-performing features currently available.

Importantly, while Neuronpedia typically displays the highest positive and negative logits for each feature, our analysis focuses on the highest positive \emph{next-token} logits (i.e., the logits predicted for the token immediately following the active feature). This distinction stems from our dataset's construction and aligns more accurately with our primary task of property-generation.

Representative features from each category available on Neuronpedia:

\begin{itemize}
    \item \textbf{Sensory ($r = 0.575$):} Feature 193, Layer 22.\footnote{\url{https://www.neuronpedia.org/gemma-3-4b-it/22-gemmascope-2-res-16k/193}} \\
    \textit{Auto-interpretation:} Edible foods and gustatory/oral content.
    \item \textbf{Motion-Self ($r = 0.426$):} Feature 132, Layer 29.\footnote{\url{https://www.neuronpedia.org/gemma-3-4b-it/29-gemmascope-2-res-16k/132}}\\
    \textit{Auto-interpretation:} Forceful or momentous events and qualities.
    \item \textbf{Internal ($r = 0.647$):} Feature 472, Layer 29.\footnote{\url{https://www.neuronpedia.org/gemma-3-4b-it/29-gemmascope-2-res-16k/472}} \\
    \textit{Auto-interpretation:} Emotional and motivational states across positive and negative affect.
    \item \textbf{Social ($r = 0.375$):} Feature 166, Layer 22.\footnote{\url{https://www.neuronpedia.org/gemma-3-4b-it/22-gemmascope-2-res-16k/166}} \\
    \textit{Auto-interpretation:} Family and kinship relations.
    \item \textbf{Time ($r = 0.373$):} Feature 733, Layer 29.\footnote{\url{https://www.neuronpedia.org/gemma-3-4b-it/29-gemmascope-2-res-16k/733}} \\
    \textit{Auto-interpretation:} Value-laden outcomes, advantages and setbacks.
\end{itemize}

\begin{figure}[t]
    \centering
    \begin{minipage}{0.8\linewidth}
        \centering
        \includegraphics[width=0.8\linewidth]{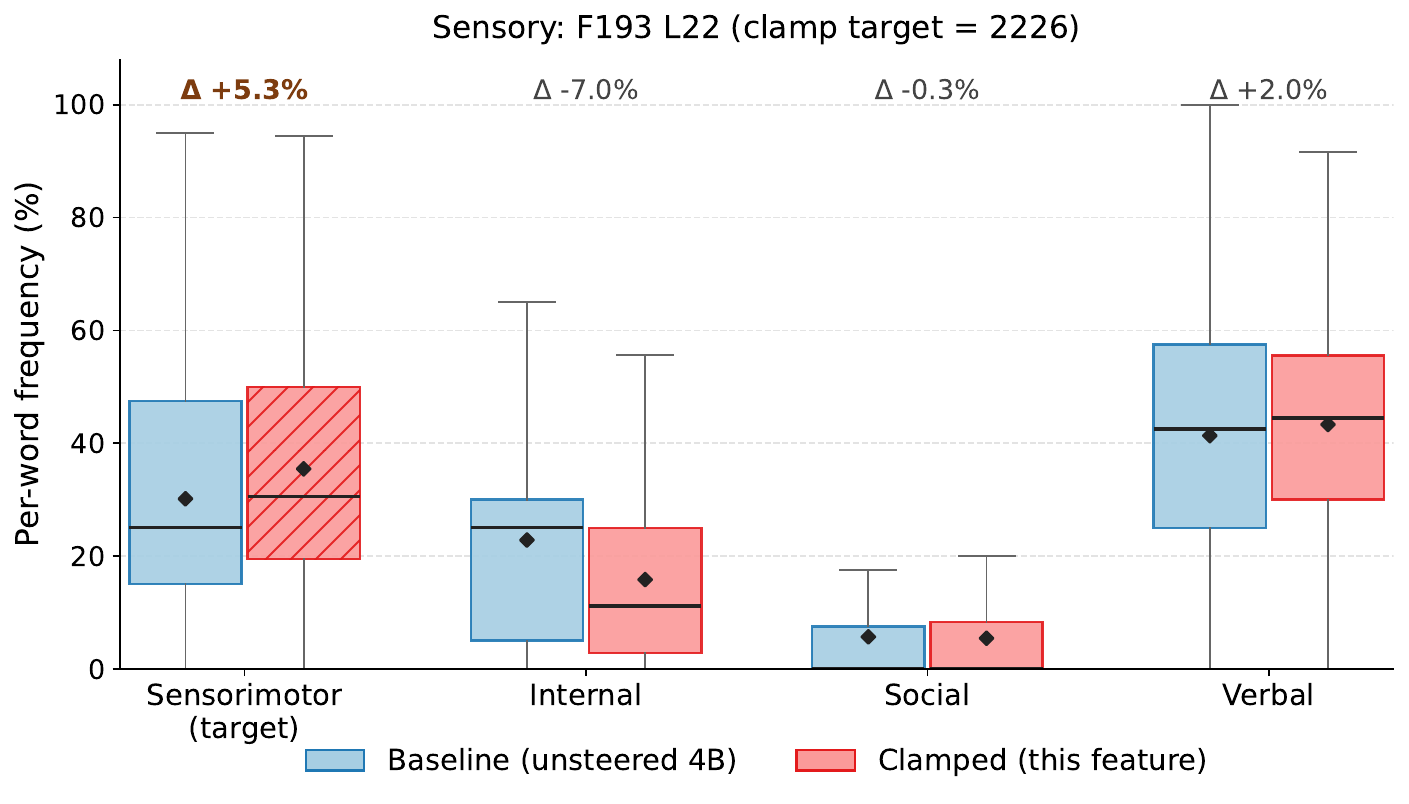}
    \end{minipage}
    
    \vspace{0.5cm} 
    
    \begin{minipage}{0.8\linewidth}
        \centering
        \includegraphics[width=0.8\linewidth]{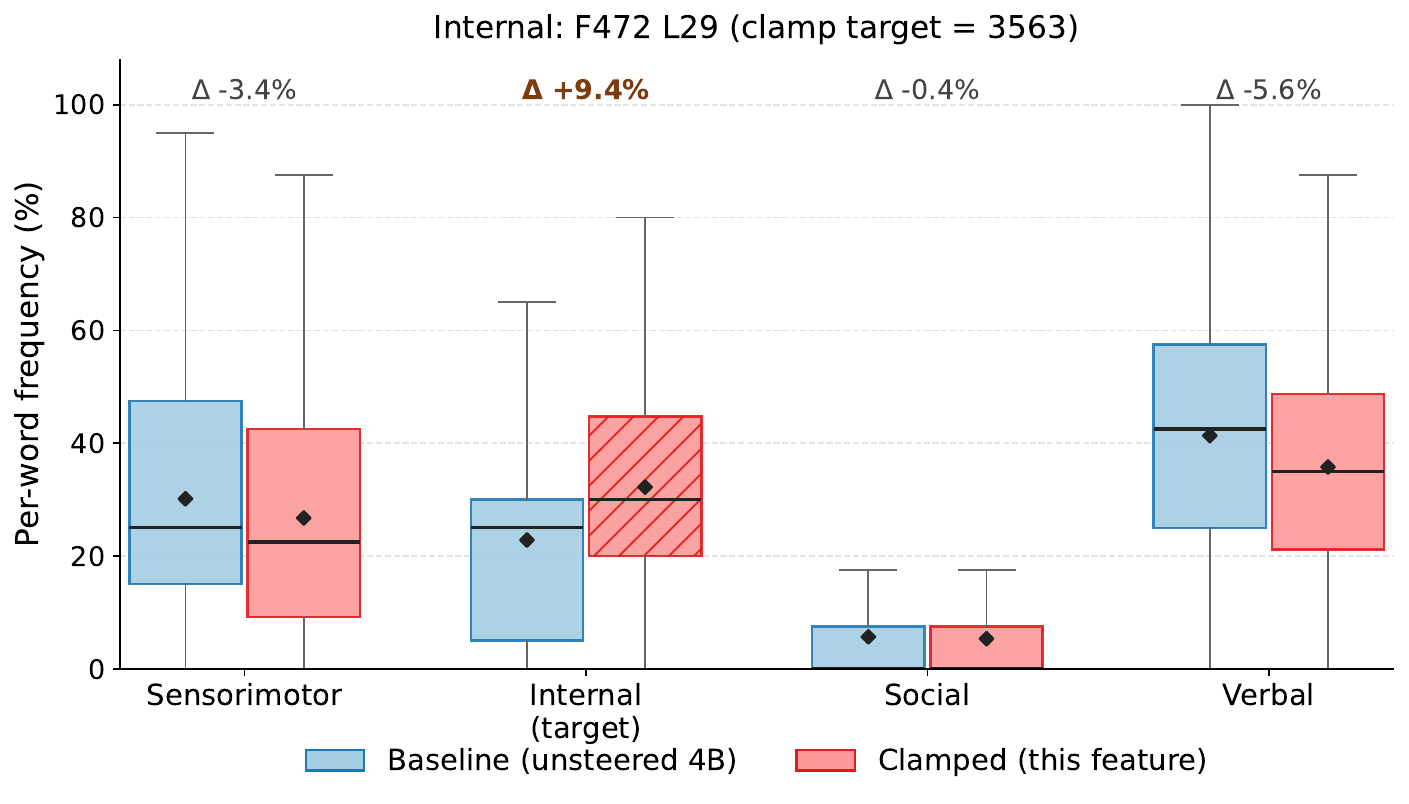}
    \end{minipage}
    
    \caption{Boxplot for steered model with sensory (top) and internal (bottom) features.}
    \label{fig:steer-boxplot}
\end{figure}

\paragraph{Feature steering.}
For steering, we use clamping \citep{templeton2024scaling} and we set the activation to double the median of the highest-activating noun from our identification dataset (Section~\ref{sec:feature-identify}). While this heuristic ensures a clear signal, it yields inconsistent intervention strengths. Specifically for the internal-feature, we found it had unusually high max-activation and we kept it at 1x, because otherwise the model did not follow instructions properly.

In Figure~\ref{fig:steer-boxplot} we present the boxplots for the Sensory and Internal features from above, compared against the baseline model. Both features lead to substantial shifts of the distribution of generated properties towards their target category, while the other categories decline or remain almost constant.

{
\section{Controlled comparison of rating and generation}
\label{app:matched}

This appendix details the controls summarized in Table~\ref{tab:matched}. All values are means over the 21 models. Ceilings are human split-half reliabilities, projected to the full panel with Spearman-Brown and recomputed on the exact word set and dimensions of each row.

\begin{table}[ht]
  \centering
  \small
  \caption{Rating vs.\ generation under matched conditions, averaged over the 21 models. Ceiling: human split-half reliability on the same words and dimensions.}
  \label{tab:matched}
  \begin{tabular}{llccc}
    \toprule
    Task & Condition & Mean $r$ & Ceiling & $r$ / ceiling \\
    \midrule
    Rating & all 751 words, 14 dimensions & \fix{0.631} & 0.871 & \fix{72.5\%} \\
    Rating & \fix{378} most abstract words, 14 dimensions & \fix{0.543} & \fix{0.751} & \fix{72.4\%} \\
    Rating & all 751 words, 3 factors & \fix{0.636} & \fix{0.917} & \fix{69.4\%} \\
    \midrule
    Rating & \fix{93} shared words, 3 shared categories & \fix{0.578} & 0.738 & \fix{78.4\%} \\
    Generation & \fix{93} shared words, 3 shared categories & \fix{0.308} & 0.966 & \fix{31.8\%} \\
    \bottomrule
  \end{tabular}
\end{table}

\paragraph{Abstract words only.}
The rating task of \citet{troche2017} spans a concreteness continuum, while Experiment~1 uses abstract words only. We therefore restricted the rating task to the 50\% most abstract words (A50, \fix{378} words, by the concreteness norms released with the dataset). As a control for the smaller sample, we repeated the analysis on 20 random draws of 376 words.

\begin{center}
\small
\begin{tabular}{lccc}
\toprule
Split & Mean $r$ (14 dimensions) & Ceiling & $r$ / ceiling \\
\midrule
Full (751 words) & \fix{0.631} & 0.871 & \fix{72.5\%} \\
Random 50\% (control, 20 draws) & \fix{0.630} & 0.870 & \fix{72.5\%} \\
A50 (\fix{378} most abstract) & \fix{0.543} & \fix{0.751} & \fix{72.4\%} \\
\bottomrule
\end{tabular}
\end{center}

The decrease on abstract words comes from the restricted range of the abstract subset, not from the smaller sample: the human ceiling drops in the same proportion, and the fraction of attainable reliability that models reach is unchanged.

\paragraph{Three-factor reduction.}
\citet{troche2017} reduce the 14 dimensions to three factors: Endogenous (affective and internal), Exogenous (perceptual and sensory), and Magnitude. Projecting human and model ratings onto these factors and correlating them per factor gives a mean $r$ of \fix{$0.636$}, compared with \fix{$0.631$} for the 14 per-dimension correlations. On the A50 subset, the three-factor correlation drops only from \fix{$0.636$} to \fix{$0.614$}, compared with \fix{$0.631$} to \fix{$0.543$} in the 14-dimension analysis; the larger drop in the latter is concentrated in individual sensory dimensions.

\paragraph{Matched words and categories.}
No rating dimension corresponds to the \textit{Verbal Association} category of Experiment~1, and the Magnitude dimensions (Space, Quantity, Time) have no counterpart among its categories. We therefore compared the two tasks on the three categories they share: Sensorimotor (Color, Taste/Smell, Tactile, Visual Form, Auditory, Self-Motion), Internal (Emotion, Polarity, Morality, Thought), and Social (Social), restricted to the \fix{93} words present in both stimulus sets.

\begin{center}
\small
\begin{tabular}{lccc}
\toprule
& Mean $r$ (3 categories) & Ceiling & $r$ / ceiling \\
\midrule
Rating (Troche) & \fix{0.578} & 0.738 & \fix{78.4\%} \\
Generation (Harpaintner) & \fix{0.308} & 0.966 & \fix{31.8\%} \\
\midrule
Difference & \fix{$+0.271$} & $-0.228$ & \fix{$+46.6$} pp \\
\bottomrule
\end{tabular}
\end{center}

\section{Alternative alignment metrics}
\label{app:metrics}

We tested whether the contrast between the two tasks depends on Pearson correlation by adding Spearman correlation and representational similarity analysis (RSA). For RSA, we build a concept-by-concept dissimilarity matrix from the human category vectors and another from each model's vectors, and correlate the two matrices. RSA therefore asks whether humans and models organize concepts in the same way, rather than whether they assign the same frequency to each category.

\begin{center}
\small
\begin{tabular}{lccc}
\toprule
\multicolumn{4}{l}{\textit{Matched setup: \fix{93} shared words, 3 shared categories}} \\
Metric & Rating & Generation & Gap \\
\midrule
Pearson & \fix{0.578} & \fix{0.308} & \fix{$+0.271$} \\
Spearman & \fix{0.506} & \fix{0.321} & \fix{$+0.185$} \\
RSA & \fix{0.491} & \fix{0.104} & \fix{$+0.387$} \\
\midrule
RSA, Claude Opus 4.6 & \fix{0.538} & \fix{0.132} & \fix{$+0.406$} \\
RSA, GPT-5.4 & \fix{0.502} & \fix{0.169} & \fix{$+0.333$} \\
RSA, Gemini 3.1 Pro & 0.529 & \fix{0.106} & \fix{$+0.423$} \\
\bottomrule
\end{tabular}
\end{center}

The contrast also holds when each task is scored on its own complete word set. Values are means over the 21 models with 95\% bootstrap confidence intervals over words (1000 draws).

\begin{center}
\small
\begin{tabular}{llccc}
\toprule
Task & Metric & Words & Mean & 95\% CI \\
\midrule
Rating & Pearson & 751 & \fix{0.631} & [0.620, 0.641] \\
Rating & Spearman & 751 & \fix{0.591} & [0.582, 0.602] \\
Rating & RSA & 751 & \fix{0.675} & [0.651, 0.698] \\
Generation & Pearson & 293 & \fix{0.281} & \fix{[0.236, 0.324]} \\
Generation & Spearman & 293 & \fix{0.279} & \fix{[0.234, 0.321]} \\
Generation & RSA & 293 & \fix{0.082} & \fix{[0.065, 0.123]} \\
\bottomrule
\end{tabular}
\end{center}

The confidence intervals of the two tasks are disjoint under every metric. The low RSA in generation formalizes the observation of Section~\ref{sec:exp1}: models can produce roughly human-like aggregate frequencies while assigning properties to different concepts than humans do.

\section{Property budget}
\label{app:budget}

The four-property budget of Experiment~1 might lead models to spend their responses on easily accessible verbal associations, pushing internal-state content past the cutoff. We repeated Experiment~1 with the three frontier models, requesting 4, 8, or 12 properties per concept (10 runs each). For 8 and 12 properties we score both the full list (\textsc{full}) and a random subset of four properties (\textsc{rand4}). SM, IS, SOC, VA denote Sensorimotor, Internal State \& Emotion, Social, and Verbal Association; humans produce $32.5\%$ IS and $26.0\%$ VA.

\begin{center}
\small
\setlength{\tabcolsep}{4pt}
\begin{tabular}{llccccc|cccc}
\toprule
Model & Condition & Mean $r$ & SM\% & IS\% & SOC\% & VA\% & SM $r$ & IS $r$ & SOC $r$ & VA $r$ \\
\midrule
GPT-5.4 & 4 (baseline) & 0.303 & 31.1 & 13.5 & 10.5 & 44.7 & 0.33 & 0.26 & 0.41 & 0.21 \\
 & 8 \textsc{full} & 0.315 & 35.4 & 14.1 & 11.6 & 38.9 & 0.34 & 0.32 & 0.40 & 0.19 \\
 & 8 \textsc{rand4} & 0.321 & 35.4 & 13.9 & 11.7 & 39.0 & 0.34 & 0.35 & 0.38 & 0.21 \\
 & 12 \textsc{full} & 0.320 & 38.3 & 13.5 & 13.2 & 34.9 & 0.35 & 0.34 & 0.41 & 0.18 \\
 & 12 \textsc{rand4} & 0.301 & 38.6 & 13.6 & 13.4 & 34.4 & 0.30 & 0.32 & 0.39 & 0.19 \\
\midrule
Claude Opus 4.6 & 4 (baseline) & 0.266 & 30.4 & 13.5 & 7.8 & 48.1 & 0.21 & 0.20 & 0.37 & 0.28 \\
 & 8 \textsc{full} & 0.337 & 30.9 & 16.0 & 10.1 & 43.0 & 0.31 & 0.36 & 0.42 & 0.25 \\
 & 8 \textsc{rand4} & 0.281 & 30.5 & 16.0 & 10.0 & 43.5 & 0.25 & 0.30 & 0.38 & 0.20 \\
 & 12 \textsc{full} & 0.352 & 33.3 & 16.5 & 11.1 & 39.1 & 0.34 & 0.40 & 0.42 & 0.24 \\
 & 12 \textsc{rand4} & 0.331 & 33.2 & 16.6 & 11.1 & 39.1 & 0.33 & 0.37 & 0.39 & 0.24 \\
\midrule
Gemini 3.1 Pro & 4 (baseline) & 0.259 & 35.2 & 7.6 & 9.1 & 47.9 & 0.31 & 0.16 & 0.40 & 0.17 \\
 & 8 \textsc{full} & 0.236 & 41.0 & 7.5 & 10.7 & 40.8 & 0.34 & 0.12 & 0.40 & 0.09 \\
 & 8 \textsc{rand4} & 0.224 & 40.7 & 7.9 & 10.8 & 40.6 & 0.32 & 0.12 & 0.35 & 0.10 \\
 & 12 \textsc{full} & 0.268 & 43.5 & 8.0 & 12.5 & 36.0 & 0.34 & 0.19 & 0.42 & 0.12 \\
 & 12 \textsc{rand4} & 0.208 & 43.1 & 8.0 & 13.0 & 35.8 & 0.29 & 0.14 & 0.34 & 0.06 \\
\bottomrule
\end{tabular}
\end{center}

A larger budget slightly lowers the share of verbal associations and raises sensorimotor and social content, but internal-state content stays at \fix{about} half the human rate or less, and Mean~$r$ stays far below the human ceiling.

\section{Prompt framing}
\label{app:framing}

We compared the original neutral instruction with two variants, each inserting one clause at the same position of the original prompt (Figure~\ref{fig:appendix-prompt-harpaintner}), so that the task description, output format, and worked example are identical across conditions.

\begin{itemize}
    \item \textit{First-person experiential:} ``As you do this, recall specific moments when you personally experienced or encountered `\{word\}', and let the properties come from those memories.''
    \item \textit{Free-associate as a person:} ``Do not aim to be complete or accurate; simply write the first things an ordinary person would associate with the word.''
\end{itemize}

\begin{center}
\small
\begin{tabular}{lccc}
\toprule
Framing & GPT-5.4 & Claude Opus 4.6 & Gemini 3.1 Pro \\
\midrule
Neutral (original) & 0.303 & 0.266 & 0.259 \\
First-person experiential & 0.242 & 0.212 & 0.159 \\
Free-associate as a person & 0.306 & 0.343 & 0.302 \\
\bottomrule
\end{tabular}
\end{center}

The experiential framing lowers alignment for all three models. The free-association framing yields modest gains for Claude Opus 4.6 and Gemini 3.1 Pro and none for GPT-5.4. The best value under either framing, $0.343$, remains far below the human ceiling, so the gap is not explained by the neutral wording of the original prompt or by models treating the task as a request for comprehensive or correct properties.

\section{Base versus instruction-tuned models}
\label{app:base}

Base models cannot follow the chat-style instruction of Experiment~1; with the original one-shot prompt in completion format, more than half of their outputs could not be parsed as four properties. We therefore used a three-shot completion prompt: the original instruction verbatim, followed by

\begin{center}
\small
\begin{verbatim}
word: sympathy
properties: hug, friends, joy, sun

word: selection
properties: wine, choice, variety, quality

word: stability
properties: balance, rock, foundation, calm

word: {word}
properties:
\end{verbatim}
\end{center}

The first example is the original one; the other two words were drawn at random from the stimulus set. The same prompt was given to the base and the instruction-tuned checkpoints. Gemma~3 12B and 27B base checkpoints did not produce reliably parsable output even under this prompt, so we report Gemma~3 4B and Qwen3 8B. Pretrained-only checkpoints are also unavailable for the frontier models we study.

\begin{center}
\small
\begin{tabular}{llcccc}
\toprule
Model & Format & Examples & Mean $r$ & Internal state (\%) & Verbal (\%) \\
\midrule
Humans (norms) & n/a & n/a & 0.974$^{\dagger}$ & 32.5 & 26.0 \\
Gemma-3-4B-it & chat & 1 & 0.270 & 22.7 & 41.4 \\
Gemma-3-4B-it & chat & 3 & 0.309 & 20.9 & 43.1 \\
Gemma-3-4B-it & completion & 3 & 0.369 & 17.1 & 41.9 \\
Gemma-3-4B-pt (base) & completion & 3 & 0.411 & 23.9 & 42.1 \\
\midrule
Qwen3-8B-it & chat & 1 & 0.373 & 15.9 & 39.6 \\
Qwen3-8B-it & chat & 3 & 0.286 & 17.7 & 45.8 \\
Qwen3-8B-it & completion & 3 & 0.359 & 17.6 & 44.0 \\
Qwen3-8B-Base & completion & 3 & \fix{0.453} & 22.5 & 48.9 \\
\bottomrule
\end{tabular}
\par\smallskip
{\footnotesize $^{\dagger}$Human ceiling.}
\end{center}

Under the matched completion format, both base checkpoints are more aligned than their instruction-tuned counterparts and produce more internal-state content, moving toward the human rate of $32.5\%$. Verbal associations, however, remain strongly over-produced, and are highest for the Qwen base model. This comparison is not a clean causal ablation: when the two additional examples were chosen to emphasize one grounding category, base-model generations shifted toward that category by about 6 to 8 percentage points, compared with about 3 to 4 points for the instruction-tuned models, so the demonstrations needed to make base models follow the task can themselves prime the category distribution. Instruction tuning may therefore influence the grounding profile, but it does not account for the gap.

\section{Statistical units and pooled-model profiles}
\label{app:units}

In \citet{harpaintner2018semantic}, each word received four properties from each of 10 participants; our model profiles likewise aggregate 10 independently sampled response sets of four properties per word. The number of response sets per word is therefore matched; what differs is that human response sets come from distinct people while model response sets are samples from one model. The diagonal of Figure~\ref{fig:inter_model_heatmap_harpaintner} shows that repeated samples from one model form a profile about as stable as the human aggregate.

To test a more heterogeneous notion of the generating ``individual'', we built pooled profiles in which each word's response sets are drawn from several models rather than one. \fix{Pooling all 21 models at 240 properties per word gives Mean~$r = 0.391$ ($40.2\%$ of the ceiling).} Even pooled profiles remain far below the human ceiling. This analysis asks a different question from our main comparison, whether heterogeneity across models produces a more human-like profile, and does not replace the model-specific results.

\section{Multimodal models}
\label{app:multimodal}

A clean test of multimodality is not possible with current frontier models, which are natively multimodal without text-only counterparts. The closest pairs in our panel are Qwen3 vs.\ Qwen3-VL at 4B and 8B. These pairs may also differ in training data, objectives, and post-training, so the comparison is suggestive rather than causal.

\begin{center}
\small
\begin{tabular}{lcc}
\toprule
Pair (VL $-$ text) & $\Delta$ Mean $r$ & $\Delta$ Sensorimotor $r$ \\
\midrule
Qwen3-4B vs.\ Qwen3-VL-4B & \fix{$-0.019$} & \fix{$+0.009$} \\
Qwen3-8B vs.\ Qwen3-VL-8B & \fix{$-0.100$} & \fix{$-0.038$} \\
\bottomrule
\end{tabular}
\end{center}

In the rating experiment, the text-only model is better in both matched pairs (Qwen3-8B $0.703$ vs.\ VL $0.517$; Qwen3-4B $0.507$ vs.\ VL $0.457$). Multimodal capability by itself therefore does not narrow the grounding gap in our data.
}

\newpage

\input{checklist.tex}

\end{document}

%% file: checklist.tex
\section*{NeurIPS Paper Checklist}

\begin{enumerate}

\item {\bf Claims}
    \item[] Question: Do the main claims made in the abstract and introduction accurately reflect the paper's contributions and scope?
    \item[] Answer: \answerYes{}
    \item[] Justification: The abstract and introduction accurately summarize the paper’s claims, experiments, and scope.

\item {\bf Limitations}
    \item[] Question: Does the paper discuss the limitations of the work performed by the authors?
    \item[] Answer: \answerYes{} 
    \item[] Justification: The paper discusses key limitations under section 6, including replication experiments, LLM-coder bias, and SAE coverage.

\item {\bf Theory assumptions and proofs}
    \item[] Question: For each theoretical result, does the paper provide the full set of assumptions and a complete (and correct) proof?
    \item[] Answer: \answerNA{} 
    \item[] Justification: The paper does not present formal theoretical results or proofs.

    \item {\bf Experimental result reproducibility}
    \item[] Question: Does the paper fully disclose all the information needed to reproduce the main experimental results of the paper to the extent that it affects the main claims and/or conclusions of the paper (regardless of whether the code and data are provided or not)?
    \item[] Answer: \answerYes{} 
    \item[] Justification: The paper provides datasets, prompts, model details, coding procedures, metrics, and uncertainty methods needed to reproduce the main results.

\item {\bf Open access to data and code}
    \item[] Question: Does the paper provide open access to the data and code, with sufficient instructions to faithfully reproduce the main experimental results, as described in supplemental material?
    \item[] Answer: \answerYes{} 
    \item[] Justification: Code, prompts, model outputs, and analysis scripts are publicly available at \url{https://github.com/odychlapanis/grounding-gap/}, with per-experiment instructions for reproducing the main results.

\item {\bf Experimental setting/details}
    \item[] Question: Does the paper specify all the training and test details (e.g., data splits, hyperparameters, how they were chosen, type of optimizer) necessary to understand the results?
    \item[] Answer: \answerYes{} 
    \item[] Justification: The experimental setup specifies stimuli, prompts, models, runs, coding, temperatures, and evaluation metrics.

\item {\bf Experiment statistical significance}
    \item[] Question: Does the paper report error bars suitably and correctly defined or other appropriate information about the statistical significance of the experiments?
    \item[] Answer: \answerYes{} 
    \item[] Justification: The paper reports bootstrap uncertainty, human ceilings, coder-error margins, and convergence analyses.

\item {\bf Experiments compute resources}
    \item[] Question: For each experiment, does the paper provide sufficient information on the computer resources (type of compute workers, memory, time of execution) needed to reproduce the experiments?
    \item[] Answer: \answerYes{} 
    \item[] Justification: The paper identifies the compute setup, including API use for closed models and 8 x Quadro RTX 6000 runs for open models.
    
\item {\bf Code of ethics}
    \item[] Question: Does the research conducted in the paper conform, in every respect, with the NeurIPS Code of Ethics \url{https://neurips.cc/public/EthicsGuidelines}?
    \item[] Answer: \answerYes{} 
    \item[] Justification: The work uses existing datasets and model outputs, preserves anonymity, and presents no apparent ethics-policy violation.

\item {\bf Broader impacts}
    \item[] Question: Does the paper discuss both potential positive societal impacts and negative societal impacts of the work performed?
    \item[] Answer: \answerYes{} 
    \item[] Justification: Although the paper does not have a significant societal impact, it discusses why grounding differences matter as LLMs are used in more human-facing and agentic roles.
    
\item {\bf Safeguards}
    \item[] Question: Does the paper describe safeguards that have been put in place for responsible release of data or models that have a high risk for misuse (e.g., pre-trained language models, image generators, or scraped datasets)?
    \item[] Answer: \answerNA{} 
    \item[] Justification: The paper does not release a high-risk model, scraped dataset, or artifact requiring misuse safeguards.

\item {\bf Licenses for existing assets}
    \item[] Question: Are the creators or original owners of assets (e.g., code, data, models), used in the paper, properly credited and are the license and terms of use explicitly mentioned and properly respected?
    \item[] Answer: \answerYes{} 
    \item[] Justification: Existing datasets, models, SAE resources, and tools are credited through citations.

\item {\bf New assets}
    \item[] Question: Are new assets introduced in the paper well documented and is the documentation provided alongside the assets?
    \item[] Answer: \answerNA{} 
    \item[] Justification: The paper does not introduce a standalone new dataset, model, or benchmark as its main contribution.

\item {\bf Crowdsourcing and research with human subjects}
    \item[] Question: For crowdsourcing experiments and research with human subjects, does the paper include the full text of instructions given to participants and screenshots, if applicable, as well as details about compensation (if any)? 
    \item[] Answer: \answerNA{} 
    \item[] Justification: The paper does not conduct new crowdsourcing or human-subject experiments.

\item {\bf Institutional review board (IRB) approvals or equivalent for research with human subjects}
    \item[] Question: Does the paper describe potential risks incurred by study participants, whether such risks were disclosed to the subjects, and whether Institutional Review Board (IRB) approvals (or an equivalent approval/review based on the requirements of your country or institution) were obtained?
    \item[] Answer: \answerNA{} 
    \item[] Justification: The paper does not conduct new crowdsourcing or human-subject experiments.

\item {\bf Declaration of LLM usage}
    \item[] Question: Does the paper describe the usage of LLMs if it is an important, original, or non-standard component of the core methods in this research? Note that if the LLM is used only for writing, editing, or formatting purposes and does \emph{not} impact the core methodology, scientific rigor, or originality of the research, declaration is not required.
    \item[] Answer: \answerYes{} 
    \item[] Justification: LLMs are a core component of the method: they code generated properties into grounding categories (Section~\ref{sec:exp-setup} and Appendix~\ref{app:coding}), generate the cloze sentences for SAE feature identification, and produce feature auto-interpretations (Appendix~\ref{app:mech-analysis}). Each use is described and validated in the paper. LLMs were also used for text editing.

\end{enumerate}

%% file: references.bib
@article{borghi2017challenge,
  author  = {Borghi, Anna M. and Binkofski, Ferdinand and Castelfranchi, Cristiano and Cimatti, Felice and Scorolli, Claudia and Tummolini, Luca},
  title   = {The Challenge of Abstract Concepts},
  journal = {Psychological Bulletin},
  year    = {2017},
  volume  = {143},
  number  = {3},
  pages   = {263--292},
  doi     = {10.1037/bul0000089}
}

@article{troche2017,
  title     = {Defining a conceptual topography of word concreteness: Clustering properties of emotion, sensation, and magnitude among 750 {E}nglish words},
  author    = {Troche, Joshua and Crutch, Sebastian J. and Reilly, Jamie},
  journal   = {Frontiers in Psychology},
  volume    = {8},
  pages     = {1787},
  year      = {2017},
  publisher = {Frontiers},
  doi       = {10.3389/fpsyg.2017.01787}
}

@article{mcrae2005semantic,
  author    = {McRae, Ken and Cree, George S. and Seidenberg, Mark S. and McNorgan, Chris},
  title     = {Semantic feature production norms for a large set of living and nonliving things},
  journal   = {Behavior Research Methods},
  volume    = {37},
  number    = {4},
  pages     = {547--559},
  year      = {2005},
  doi       = {10.3758/BF03192726}
}

@article{xu2025large,
  author    = {Xu, Qihui and Peng, Yingying and Nastase, Samuel A. and Chodorow, Martin and Wu, Minghua and Li, Ping},
  title     = {Large language models without grounding recover non-sensorimotor but not sensorimotor features of human concepts},
  journal   = {Nature Human Behaviour},
  volume    = {9},
  pages     = {1871--1886},
  year      = {2025},
  doi       = {10.1038/s41562-025-02203-8}
}

@article{barsalou1999perceptual,
  title     = {Perceptual symbol systems},
  author    = {Barsalou, Lawrence W.},
  journal   = {Behavioral and Brain Sciences},
  volume    = {22},
  number    = {4},
  pages     = {577--660},
  year      = {1999},
  publisher = {Cambridge University Press},
  doi       = {10.1017/S0140525X99002149}
}

@incollection{barsalou2005situating,
  title     = {Situating abstract concepts},
  author    = {Barsalou, Lawrence W. and Wiemer-Hastings, Katinka},
  booktitle = {Grounding Cognition: The Role of Perception and Action in Memory, Language, and Thought},
  editor    = {Pecher, Diane and Zwaan, Rolf A.},
  pages     = {129--163},
  year      = {2005},
  publisher = {Cambridge University Press},
  doi       = {10.1017/CBO9780511499968.007}
}

@book{lakoff1980metaphors,
 title={Metaphors We Live By}, ISBN={9780226470993}, url={http://dx.doi.org/10.7208/chicago/9780226470993.001.0001}, DOI={10.7208/chicago/9780226470993.001.0001}, publisher={University of Chicago Press}, author={Lakoff, George and Johnson, Mark}, year={2003} 
 }

@article{harpaintner2018semantic,
  title     = {The semantic content of abstract concepts: {A} property listing study of 296 abstract words},
  author    = {Harpaintner, Markus and Trumpp, Natalie M. and Kiefer, Markus},
  journal   = {Frontiers in Psychology},
  volume    = {9},
  pages     = {1748},
  year      = {2018},
  publisher = {Frontiers Media},
  doi       = {10.3389/fpsyg.2018.01748}
}

@article{kelly2024conceptual,
  author  = {Kelly, Aubrey E. and Kenett, Yoed N. and Medaglia, John D. and Reilly, Jamie J. and Dudhat, Priyanka and Chrysikou, Evangelia G.},
  title   = {Conceptual structure of emotions},
  journal = {Emotion},
  year    = {2024},
  volume  = {24},
  number  = {6},
  pages   = {1550--1561},
  doi     = {10.1037/emo0001327}
}

@article{brysbaert2014concreteness,
  title     = {Concreteness ratings for 40 thousand generally known {English} word lemmas},
  author    = {Brysbaert, Marc and Warriner, Amy Beth and Kuperman, Victor},
  journal   = {Behavior Research Methods},
  volume    = {46},
  number    = {3},
  pages     = {904--911},
  year      = {2014},
  publisher = {Springer},
  doi       = {10.3758/s13428-013-0403-5}
}

@article{lynott2020lancaster,
  title     = {The {Lancaster} Sensorimotor Norms: multidimensional measures of perceptual and action strength for 40,000 {English} words},
  author    = {Lynott, Dermot and Connell, Louise and Brysbaert, Marc and Brand, James and Carney, James},
  journal   = {Behavior Research Methods},
  volume    = {52},
  pages     = {1271--1291},
  year      = {2020},
  publisher = {Springer},
  doi       = {10.3758/s13428-019-01316-z}
}

@article{scott2019glasgow,
  title     = {The {Glasgow} Norms: ratings of 5,500 words on nine scales},
  author    = {Scott, Graham G. and Keitel, Anne and Becirspahic, Mia and Yao, Bo and Sereno, Sara C.},
  journal   = {Behavior Research Methods},
  volume    = {51},
  pages     = {1258--1270},
  year      = {2019},
  publisher = {Springer},
  doi       = {10.3758/s13428-018-1099-3}
}

@article{diveica2023quantifying,
  title     = {Quantifying social semantics: An inclusive definition of socialness and ratings for 8,388 {English} words},
  author    = {Diveica, Veronica and Pexman, Penny M. and Binney, Richard J.},
  journal   = {Behavior Research Methods},
  volume    = {55},
  pages     = {461--473},
  year      = {2023},
  publisher = {Springer},
  doi       = {10.3758/s13428-022-01810-x}
}

@inproceedings{suresh2023conceptual,
  title     = {Conceptual Structure Coheres in Human Cognition but Not in Large Language Models},
  author    = {Suresh, Siddharth and Mukherjee, Kushin and Yu, Xizheng and Huang, Wei-Chun and Padua, Lisa and Rogers, Timothy T.},
  booktitle = {Proceedings of the 2023 Conference on Empirical Methods in Natural Language Processing (EMNLP)},
  pages     = {722--738},
  year      = {2023},
  url       = {https://aclanthology.org/2023.emnlp-main.47/}
}

@article{pezzelle2021word,
  title   = {Word Representation Learning in Multimodal Pre-Trained Transformers: An Intrinsic Evaluation},
  author  = {Pezzelle, Sandro and Takmaz, Ece and Fern{\'a}ndez, Raquel},
  journal = {Transactions of the Association for Computational Linguistics (TACL)},
  volume  = {9},
  pages   = {1563--1579},
  year    = {2021},
  url     = {https://aclanthology.org/2021.tacl-1.93/}
}

@misc{wang2025cognitive,
  title  = {Cognitive Alignment Between Humans and LLMs Across Multimodal Domains},
  author = {Wang, Yuwei and Liang, Dongqi and Zeng, Yi},
  year   = {2025},
  month  = jan,
  note   = {Research Square preprint},
  doi    = {10.21203/rs.3.rs-5736241/v1},
  url    = {https://www.researchsquare.com/article/rs-5736241/v1}
}

@article{trott2024can,
title={Can large language models help augment English psycholinguistic datasets?}, volume={56}, 
 url={http://dx.doi.org/10.3758/s13428-024-02337-z}, 
 DOI={10.3758/s13428-024-02337-z}, 
 number={6}, journal={Behavior Research Methods}, publisher={Springer Science and Business Media LLC}, author={Trott, Sean}, year={2024}, month=Jan, pages={6082–6100} 
 }

@inproceedings{abdou2021language,
  title     = {Can Language Models Encode Perceptual Structure Without Grounding? A Case Study in Color},
  author    = {Abdou, Mostafa and Kulmizev, Artur and Hershcovich, Daniel and Frank, Stella and Pavlick, Ellie and S{\o}gaard, Anders},
  booktitle = {Proceedings of the 25th Conference on Computational Natural Language Learning (CoNLL)},
  year      = {2021},
  url       = {https://aclanthology.org/2021.conll-1.9/}
}

@inproceedings{gurnee2024language,
 author = {Gurnee, Wes and Tegmark, Max },
 booktitle = {International Conference on Learning Representations},
 editor = {B. Kim and Y. Yue and S. Chaudhuri and K. Fragkiadaki and M. Khan and Y. Sun},
 pages = {2483--2503},
 title = {Language Models Represent Space and Time},
 url = {https://proceedings.iclr.cc/paper_files/paper/2024/file/0a6059857ae5c82ea9726ee9282a7145-Paper-Conference.pdf},
 volume = {2024},
 year = {2024}
}

@inproceedings{tigges2023linear,
    title = "Language Models Linearly Represent Sentiment",
    author = "Tigges, Curt  and
      Hollinsworth, Oskar J.  and
      Geiger, Atticus  and
      Nanda, Neel",
    editor = "Belinkov, Yonatan  and
      Kim, Najoung  and
      Jumelet, Jaap  and
      Mohebbi, Hosein  and
      Mueller, Aaron  and
      Chen, Hanjie",
    booktitle = "Proceedings of the 7th BlackboxNLP Workshop: Analyzing and Interpreting Neural Networks for NLP",
    month = nov,
    year = "2024",
    address = "Miami, Florida, US",
    publisher = "Association for Computational Linguistics",
    url = "https://aclanthology.org/2024.blackboxnlp-1.5/",
    doi = "10.18653/v1/2024.blackboxnlp-1.5",
    pages = "58--87"
}

@inproceedings{li2023emergent,
title={Emergent World Representations: Exploring a Sequence Model Trained on a Synthetic Task},
author={Kenneth Li and Aspen K Hopkins and David Bau and Fernanda Vi{\'e}gas and Hanspeter Pfister and Martin Wattenberg},
booktitle={The Eleventh International Conference on Learning Representations },
year={2023},
url={https://openreview.net/forum?id=DeG07_TcZvT}
}

@inproceedings{nanda2023emergent,
    title = "Emergent Linear Representations in World Models of Self-Supervised Sequence Models",
    author = "Nanda, Neel  and
      Lee, Andrew  and
      Wattenberg, Martin",
    editor = "Belinkov, Yonatan  and
      Hao, Sophie  and
      Jumelet, Jaap  and
      Kim, Najoung  and
      McCarthy, Arya  and
      Mohebbi, Hosein",
    booktitle = "Proceedings of the 6th BlackboxNLP Workshop: Analyzing and Interpreting Neural Networks for NLP",
    month = dec,
    year = "2023",
    address = "Singapore",
    publisher = "Association for Computational Linguistics",
    url = "https://aclanthology.org/2023.blackboxnlp-1.2/",
    doi = "10.18653/v1/2023.blackboxnlp-1.2",
    pages = "16--30"
}

@techreport{gemmascope2_2025,
  title       = {Gemma Scope 2 - Technical Paper},
  author      = {McDougall, Callum and Conmy, Arthur and Kram{\'a}r, J{\'a}nos and Lieberum, Tom and Rajamanoharan, Senthooran and Nanda, Neel},
  institution = {Google DeepMind},
  year        = {2025},
  month       = sep,
  note        = {\url{https://storage.googleapis.com/deepmind-media/DeepMind.com/Blog/gemma-scope-2-helping-the-ai-safety-community-deepen-understanding-of-complex-language-model-behavior/Gemma_Scope_2_Technical_Paper.pdf}}
}

@misc{wu2025ai,
  title         = {AI Shares Emotion with Humans across Languages and Cultures},
  author        = {Wu, Xiuwen and Wang, Hao and Yan, Zhiang and Tang, Xiaohan and Xu, Pengfei and Siok, Wai-Ting and Li, Ping and Gao, Jia-Hong and Lyu, Bingjiang and Qin, Lang},
  year          = {2025},
  eprint        = {2506.13978},
  archivePrefix = {arXiv},
  primaryClass  = {cs.CL},
  url           = {https://arxiv.org/abs/2506.13978}
}

@misc{wang2025feel,
  title         = {Do {LLMs} ``Feel''? Emotion Circuits Discovery and Control},
  author        = {Wang, Chenxi and Zhang, Yixuan and Yu, Ruiji and Zheng, Yufei and Gao, Lang and Song, Zirui and Xu, Zixiang and Xia, Gus and Zhang, Huishuai and Zhao, Dongyan and Chen, Xiuying},
  year          = {2025},
  month         = oct,
  eprint        = {2510.11328},
  archivePrefix = {arXiv},
  primaryClass  = {cs.CL},
  url           = {https://arxiv.org/abs/2510.11328}
}

@article{Critchleyetal2004,
  author  = {Critchley, Hugo D. and Wiens, Stefan and Rotshtein, Pia and Öhman, Arne and Dolan, Raymond J.},
  title   = {Neural systems supporting interoceptive awareness},
  journal = {Nature Neuroscience},
  year    = {2004},
  volume  = {7},
  number  = {2},
  pages   = {189--195},
  doi     = {10.1038/nn1176}
}

@article{ettinger2020bert,
  title     = {What {BERT} Is Not: Lessons from a New Suite of Psycholinguistic Diagnostics for Language Models},
  author    = {Ettinger, Allyson},
  journal   = {Transactions of the Association for Computational Linguistics},
  volume    = {8},
  pages     = {34--48},
  year      = {2020},
  publisher = {MIT Press},
  doi       = {10.1162/tacl_a_00298}
}

@inproceedings{cunningham2024sparse,
title={Sparse Autoencoders Find Highly Interpretable Features in Language Models},
author={Robert Huben and Hoagy Cunningham and Logan Riggs Smith and Aidan Ewart and Lee Sharkey},
booktitle={The Twelfth International Conference on Learning Representations},
year={2024},
url={https://openreview.net/forum?id=F76bwRSLeK}
}

@article{bricken2023monosemanticity,
       title={Towards Monosemanticity: Decomposing Language Models With Dictionary Learning},
       author={Bricken, Trenton and Templeton, Adly and Batson, Joshua and Chen, Brian and Jermyn, Adam and Conerly, Tom and Turner, Nick and Anil, Cem and Denison, Carson and Askell, Amanda and Lasenby, Robert and Wu, Yifan and Kravec, Shauna and Schiefer, Nicholas and Maxwell, Tim and Joseph, Nicholas and Hatfield-Dodds, Zac and Tamkin, Alex and Nguyen, Karina and McLean, Brayden and Burke, Josiah E and Hume, Tristan and Carter, Shan and Henighan, Tom and Olah, Christopher},
       year={2023},
       journal={Transformer Circuits Thread},
       note={https://transformer-circuits.pub/2023/monosemantic-features/index.html}
    }

@article{templeton2024scaling,
       title={Scaling Monosemanticity: Extracting Interpretable Features from Claude 3 Sonnet},
       author={Templeton, Adly and Conerly, Tom and Marcus, Jonathan and Lindsey, Jack and Bricken, Trenton and Chen, Brian and Pearce, Adam and Citro, Craig and Ameisen, Emmanuel and Jones, Andy and Cunningham, Hoagy and Turner, Nicholas L and McDougall, Callum and MacDiarmid, Monte and Freeman, C. Daniel and Sumers, Theodore R. and Rees, Edward and Batson, Joshua and Jermyn, Adam and Carter, Shan and Olah, Chris and Henighan, Tom},
       year={2024},
       journal={Transformer Circuits Thread},
       url={https://transformer-circuits.pub/2024/scaling-monosemanticity/index.html}
    }

@article{emergent_2022,
title={Emergent Abilities of Large Language Models},
author={Jason Wei and Yi Tay and Rishi Bommasani and Colin Raffel and Barret Zoph and Sebastian Borgeaud and Dani Yogatama and Maarten Bosma and Denny Zhou and Donald Metzler and Ed H. Chi and Tatsunori Hashimoto and Oriol Vinyals and Percy Liang and Jeff Dean and William Fedus},
journal={Transactions on Machine Learning Research},
issn={2835-8856},
year={2022},
url={https://openreview.net/forum?id=yzkSU5zdwD},
note={Survey Certification}
}

@inproceedings{lieberum-etal-2024-gemma,
    title = "Gemma Scope: Open Sparse Autoencoders Everywhere All At Once on Gemma 2",
    author = "Lieberum, Tom  and
      Rajamanoharan, Senthooran  and
      Conmy, Arthur  and
      Smith, Lewis  and
      Sonnerat, Nicolas  and
      Varma, Vikrant  and
      Kramar, Janos  and
      Dragan, Anca  and
      Shah, Rohin  and
      Nanda, Neel",
    editor = "Belinkov, Yonatan  and
      Kim, Najoung  and
      Jumelet, Jaap  and
      Mohebbi, Hosein  and
      Mueller, Aaron  and
      Chen, Hanjie",
    booktitle = "Proceedings of the 7th BlackboxNLP Workshop: Analyzing and Interpreting Neural Networks for NLP",
    month = nov,
    year = "2024",
    address = "Miami, Florida, US",
    publisher = "Association for Computational Linguistics",
    url = "https://aclanthology.org/2024.blackboxnlp-1.19/",
    doi = "10.18653/v1/2024.blackboxnlp-1.19",
    pages = "278--300"
}

@inproceedings{panickssery2024steering,
  title     = {Steering {LLaMA 2} via contrastive activation addition},
  author    = {Panickssery, Nina and Gabrieli, Nick and Schulz, Julian and Tong, Meg and Hubinger, Evan and Turner, Alex},
  booktitle = {Proceedings of the 62nd Annual Meeting of the Association for Computational Linguistics (Volume 1: Long Papers)},
  pages     = {15504--15522},
  year      = {2024},
  publisher = {Association for Computational Linguistics},
  doi       = {10.18653/v1/2024.acl-long.828}
}

@misc{turner2024activation,
      title={Steering Language Models With Activation Engineering}, 
      author={Alexander Matt Turner and Lisa Thiergart and Gavin Leech and David Udell and Juan J. Vazquez and Ulisse Mini and Monte MacDiarmid},
      year={2024},
      eprint={2308.10248},
      archivePrefix={arXiv},
      primaryClass={cs.CL},
      url={https://arxiv.org/abs/2308.10248}, 
}

@article{villani2019,
  title     = {Varieties of abstract concepts and their multiple dimensions},
  author    = {Villani, Caterina and Lugli, Luisa and Liuzza, Marco Tullio and Borghi, Anna M.},
  journal   = {Language and Cognition},
  volume    = {11},
  number    = {3},
  pages     = {403--430},
  year      = {2019},
  publisher = {Cambridge University Press},
  doi       = {10.1017/langcog.2019.23}
}

@incollection{Barsalou2026Grounded,
  author    = {Barsalou, Lawrence W.},
  title     = {Grounded Cognition},
  booktitle = {Open Encyclopedia of Cognitive Science},
  editor    = {Frank, Michael C. and Majid, Asifa},
  publisher = {MIT Press},
  year      = {2026},
  url       = {https://oecs.mit.edu/pub/9iq4376o},
}

@inproceedings{bender-koller-2020-climbing,
    title = "Climbing towards {NLU}: {On} Meaning, Form, and Understanding in the Age of Data",
    author = "Bender, Emily M.  and
      Koller, Alexander",
    editor = "Jurafsky, Dan  and
      Chai, Joyce  and
      Schluter, Natalie  and
      Tetreault, Joel",
    booktitle = "Proceedings of the 58th Annual Meeting of the Association for Computational Linguistics",
    month = jul,
    year = "2020",
    address = "Online",
    publisher = "Association for Computational Linguistics",
    url = "https://aclanthology.org/2020.acl-main.463/",
    doi = "10.18653/v1/2020.acl-main.463",
    pages = "5185--5198"
}

@inproceedings{bender2021dangers,
author = {Bender, Emily M. and Gebru, Timnit and McMillan-Major, Angelina and Shmitchell, Shmargaret},
title = {On the Dangers of Stochastic Parrots: Can Language Models Be Too Big?},
year = {2021},
isbn = {9781450383097},
publisher = {Association for Computing Machinery},
address = {New York, NY, USA},
url = {https://doi.org/10.1145/3442188.3445922},
doi = {10.1145/3442188.3445922},
booktitle = {Proceedings of the 2021 ACM Conference on Fairness, Accountability, and Transparency},
pages = {610–623},
numpages = {14},
location = {Virtual Event, Canada},
series = {FAccT '21}
}

@article{hauk2004somatotopic,
  title   = {Somatotopic representation of action words in human motor and premotor cortex},
  author  = {Hauk, Olaf and Johnsrude, Ingrid and Pulverm{\"u}ller, Friedemann},
  journal = {Neuron},
  volume  = {41},
  number  = {2},
  pages   = {301--307},
  year    = {2004},
  doi     = {10.1016/S0896-6273(03)00838-9}
}

@article{pulvermuller2005brain,
  title   = {Brain mechanisms linking language and action},
  author  = {Pulverm{\"u}ller, Friedemann},
  journal = {Nature Reviews Neuroscience},
  volume  = {6},
  number  = {7},
  pages   = {576--582},
  year    = {2005},
  doi     = {10.1038/nrn1706}
}

@misc{sofroniew2026emotions,
      title={Emotion Concepts and their Function in a Large Language Model}, 
      author={Nicholas Sofroniew and Isaac Kauvar and William Saunders and Runjin Chen and Tom Henighan and Sasha Hydrie and Craig Citro and Adam Pearce and Julius Tarng and Wes Gurnee and Joshua Batson and Sam Zimmerman and Kelley Rivoire and Kyle Fish and Chris Olah and Jack Lindsey},
      year={2026},
      eprint={2604.07729},
      archivePrefix={arXiv},
      primaryClass={cs.AI},
      url={https://arxiv.org/abs/2604.07729}, 
}

@article{mancano2026emotional,
  title   = {Emotional and Social Dimension of Abstract Concepts Meet with Interoception in Right Anterior Insula},
  author  = {Mancano, Martina and Papagno, Costanza},
  journal = {Journal of Neuroscience},
  volume  = {46},
  number  = {2},
  pages   = {e0238252025},
  year    = {2026},
  doi     = {10.1523/JNEUROSCI.0238-25.2025}
}

@article{li2026incompressible,
  title={Incompressible Knowledge Probes: Estimating Black-Box LLM Parameter Counts via Factual Capacity},
  author={Li, Bojie},
  journal={arXiv preprint arXiv:2604.24827},
  year={2026}
}

@inproceedings{NEURIPS2022_c1e2faff,
author = {Hoffmann, Jordan and Borgeaud, Sebastian and Mensch, Arthur and Buchatskaya, Elena and Cai, Trevor and Rutherford, Eliza and de Las Casas, Diego and Hendricks, Lisa Anne and Welbl, Johannes and Clark, Aidan and Hennigan, Tom and Noland, Eric and Millican, Katie and van den Driessche, George and Damoc, Bogdan and Guy, Aurelia and Osindero, Simon and Simonyan, Karen and Elsen, Erich and Vinyals, Oriol and Rae, Jack W. and Sifre, Laurent},
title = {Training compute-optimal large language models},
year = {2022},
isbn = {9781713871088},
publisher = {Curran Associates Inc.},
address = {Red Hook, NY, USA},
booktitle = {Proceedings of the 36th International Conference on Neural Information Processing Systems},
articleno = {2176},
numpages = {15},
location = {New Orleans, LA, USA},
series = {NIPS '22}
}

@misc{zheng2023judging,
 author = {Zheng, Lianmin and Chiang, Wei-Lin and Sheng, Ying and Zhuang, Siyuan and Wu, Zhanghao and Zhuang, Yonghao and Lin, Zi and Li, Zhuohan and Li, Dacheng and Xing, Eric and Zhang, Hao and Gonzalez, Joseph and Stoica, Ion},
 booktitle = {Advances in Neural Information Processing Systems},
 editor = {A. Oh and T. Naumann and A. Globerson and K. Saenko and M. Hardt and S. Levine},
 pages = {46595--46623},
 publisher = {Curran Associates, Inc.},
 title = {Judging LLM-as-a-Judge with MT-Bench and Chatbot Arena},
 url = {https://proceedings.neurips.cc/paper_files/paper/2023/file/91f18a1287b398d378ef22505bf41832-Paper-Datasets_and_Benchmarks.pdf},
 volume = {36},
 year = {2023}
}

@article{fung2025embodied,
  title={Embodied ai agents: Modeling the world},
  author={Fung, Pascale and Bachrach, Yoram and Celikyilmaz, Asli and Chaudhuri, Kamalika and Chen, Delong and Chung, Willy and Dupoux, Emmanuel and Gong, Hongyu and J{\'e}gou, Herv{\'e} and Lazaric, Alessandro and others},
  journal={arXiv preprint arXiv:2506.22355},
  year={2025}
}

@article{brooks1991intelligence,
title = {Intelligence without representation},
journal = {Artificial Intelligence},
volume = {47},
number = {1},
pages = {139-159},
year = {1991},
issn = {0004-3702},
doi = {https://doi.org/10.1016/0004-3702(91)90053-M},
url = {https://www.sciencedirect.com/science/article/pii/000437029190053M},
author = {Rodney A. Brooks}
}
